\documentclass{article}

\usepackage{microtype}
\usepackage{subfigure}
\usepackage{booktabs} 

\usepackage{hyperref}

\usepackage[accepted]{icml2023}

\usepackage{amsmath}
\usepackage{amssymb}
\usepackage{mathtools}
\usepackage{amsthm}

\usepackage[capitalize,noabbrev]{cleveref}

\theoremstyle{plain}
\newtheorem{theorem}{Theorem}[section]

\newtheorem{lemma}[theorem]{Lemma}
\newtheorem{corollary}[theorem]{Corollary}
\theoremstyle{definition}
\newtheorem{definition}[theorem]{Definition}

\theoremstyle{remark}

\usepackage[textsize=tiny]{todonotes}

\usepackage{makecell}
\usepackage[utf8]{inputenc} 
\usepackage[T1]{fontenc}    
\usepackage{hyperref}       
\usepackage{url}            
\usepackage{booktabs}       
\usepackage{amsfonts}       
\usepackage{commands}
\usepackage{nicefrac}       
\usepackage{amsmath}
\usepackage{microtype}      
\usepackage{xcolor}         
\usepackage{enumitem}
\usepackage{multirow}
\usepackage[resetlabels]{multibib}
\newcites{Main}{References}
\newcites{Appendix}{References}

\allowdisplaybreaks

\usepackage{tikz}
\usepackage{nicefrac}
\usetikzlibrary{shapes,arrows,positioning,calc}
\usetikzlibrary{bayesnet}

\tikzset{rv/.style={circle, draw, thick, minimum size=6mm, inner sep=0.5mm}}
\tikzset{rve/.style={ellipse, draw, thick, minimum size=6mm, inner sep=0.3mm}}
\tikzset{fv/.style={rectangle, draw, thick, minimum size=6mm, inner sep=0.5mm}}
\tikzset{rvl/.style={circle, red, draw, thick, minimum size=6mm, inner sep=0.3mm}}
\tikzset{rv0/.style={circle, draw, thick, minimum size=6mm, inner sep=0.3mm}}
\tikzset{rv1/.style={circle, draw, minimum size=6mm, inner sep=0.3mm}}
\tikzset{deg/.style={->, very thick}}
\tikzset{degl/.style={->, very thick, color=red}}
\tikzset{begl/.style={<->, very thick, color=red, dashed}}
\tikzset{beg/.style={<->, very thick, color=red}}

\usepackage{blindtext}
\usepackage{subcaption}
\usepackage{caption}
\newcommand{\Do}{\operatorname{do}}
\newcommand{\effect}{\Psi}

\newtheoremstyle{mystyle}
  {11pt}
  {11pt}
  {}
  {}
  {\bfseries}
  {.}
  { }
  {\thmname{#1}\thmnumber{ #2}\thmnote{ (#3)}}

\theoremstyle{mystyle}
\makeatletter
\newcommand{\newreptheorem}[2]{\newtheorem*{rep@#1}{\rep@title}\newenvironment{rep#1}[1]{\def\rep@title{#2 \ref*{##1}}\begin{rep@#1}}{\end{rep@#1}}}
\makeatother

\newtheorem{property}{Property}[section]
\newreptheorem{property}{Property}
\newreptheorem{lemma}{Lemma}
\newreptheorem{corollary}{Corollary}

\newcommand{\coalitionshapley}[4]{\phi^{#1}_{#3,#2}(#4)}
\newcommand{\estcoalitionshapley}[5]{\hat{\phi}^{#1,#2}_{#4,#3}(#5)}

\newcommand{\indep}{\perp\!\!\!\!\perp}

\makeatletter
\newcommand{\subalign}[1]{%
  \vcenter{%
    \Let@ \restore@math@cr \default@tag
    \baselineskip\fontdimen10 \scriptfont\tw@
    \advance\baselineskip\fontdimen12 \scriptfont\tw@
    \lineskip\thr@@\fontdimen8 \scriptfont\thr@@
    \lineskiplimit\lineskip
    \ialign{\hfil$\m@th\scriptstyle##$&$\m@th\scriptstyle{}##$\hfil\crcr
      #1\crcr
    }%
  }%
}
\makeatother

\icmltitlerunning{PWSHAP: A Path-Wise Explanation Model for Targeted Variables}

\begin{document}

\twocolumn[
\icmltitle{PWSHAP: A Path-Wise Explanation Model for Targeted Variables}



\icmlsetsymbol{equal}{*}

\begin{icmlauthorlist}
\icmlauthor{Lucile Ter-Minassian}{equal,oxfordstats}
\icmlauthor{Oscar Clivio}{equal,oxfordstats}
\icmlauthor{Karla Diaz-Ordaz}{uclstats}
\icmlauthor{Robin J.~Evans}{oxfordstats}
\icmlauthor{Chris Holmes}{oxfordstats,turing}
\end{icmlauthorlist}

\icmlaffiliation{oxfordstats}{Department of Statistics, University of Oxford, Oxford, UK}

\icmlaffiliation{uclstats}{Department of Statistical Science, University College London, London, UK}

\icmlaffiliation{turing}{The Alan Turing Institute, London, UK}

\icmlcorrespondingauthor{Lucile Ter-Minassian}{lucile.ter-minassian@stats.ox.ac.uk}
\icmlcorrespondingauthor{Oscar Clivio}{oscar.clivio@stats.ox.ac.uk}

\icmlkeywords{Machine Learning, ICML}

\vskip 0.3in
]



\printAffiliationsAndNotice{\icmlEqualContribution} 

\begin{abstract}
Predictive black-box models can exhibit high-accuracy but their opaque nature hinders their uptake in safety-critical deployment environments. Explanation methods (XAI) can provide confidence for decision-making through increased transparency. However, existing XAI methods are not tailored towards models in sensitive domains where one predictor is of special interest, such as a treatment effect in a clinical model, or ethnicity in policy models. We introduce Path-Wise Shapley effects (PWSHAP), a framework for assessing the targeted effect of a binary (e.g.~treatment) variable from a complex outcome model. Our approach augments the predictive model with a user-defined directed acyclic graph (DAG). The method then uses the graph alongside on-manifold Shapley values to identify effects along causal pathways whilst maintaining robustness to adversarial attacks. We establish error bounds for the identified path-wise Shapley effects and for Shapley values. We show PWSHAP can perform local bias and mediation analyses with faithfulness to the model. Further, if the targeted variable is randomised we can quantify local effect modification. We demonstrate the resolution, interpretability and true locality of our approach on examples and a real-world experiment. 
\end{abstract}

\section{Introduction}
\label{sec:intro}
Recent years have seen an increase in the demand for transparency on machine learning-based decisions. In safety-sensitive settings particularly, practitioners need to understand how a model reasons in order to ensure its safe deployment in the future. In many scenarios, their attention focuses on assessing the importance of a specific predictor variable such as a treatment in a clinical model, or ethnicity with regards to model fairness. Ultimately, as humans naturally have a \emph{causal} approach to model explainability, users may want to understand how the treatment\footnote{We use the example of ``treatment'' in clinical models as illustrative of a central binary predictor variable.} causally impacts the outcome i.e.\ through what mechanisms and if this matches their prior assumptions on the causal relationships in the data. Here, we focus on the following question: in the presence of a general black-box ML model, how can we compute feature attributions according to the causal beliefs encapsulated by the posited DAG? We provide a framework for locally explaining a treatment's effect in such settings, with the following goals: \textit{reliability}, \textit{safety}, \textit{interpretability} and \textit{high resolution}. 

\textbf{Reliability, safety, interpretability in XAI} 
An XAI model should aim at generating explanations that are \emph{reliable}, \emph{safe} and \emph{interpretable}. Reliability, also known as being ``true to the model'', implies that the explanations do reveal the functional dependence of the model and are robust to distributional shifts. Safety relates to the ability to protect the framework from hazard, in particular attempts to fool models with deceptive data, also known as adversarial attacks. As defined in \citeMain{miller2019explanation}, interpretability is the degree to which a human can understand the cause of a decision. 


\textbf{High resolution for safety-critical XAI}
In addition to the XAI goals cited above, resolution is often necessary when explaining a model to ensure its fairness. Following \citeMain{chiappa2019path}, we illustrate our point using a simple causal structure inspired by the Berkeley admissions dataset. Consider a predictive model for college entry with three features: sex, exam results and department. The sensitive attribute, sex, potentially impacts the predicted admission through both fair and unfair causal pathways. Sex may indirectly and fairly impact admission due to some individuals applying to more competitive departments. However, there may also be an effect through an unfair direct path, representing prejudice on the part of the admissions officer. This motivates \emph{path-specific} measures of feature importance, instead of an overall single score that groups all paths together. 

\newpage
\textbf{Our solution: Path-Wise Shapley (PWSHAP)} \quad 

We introduce a method for explaining the local effect of a binary treatment under an assumed causal graph. We assume that (i) the treatment is an ancestor of the outcome in the directed acyclic graph (DAG) and that (ii) the DAG is compatible with the data, i.e.\ that it respects all conditional independences that could be found by running conditional independence tests in the data.  The posited DAG may come from prior domain knowledge, or indeed be learnt from data and represents the user's beliefs. Our aim is not to understand the model's ``internal DAG'', and thus we do not assume that the posited DAG corresponds to the DAG of the underlying model.

We show how augmenting the predictive model with such a causal DAG supports a novel targeted extension of SHAP values, allowing for the decomposition of the black-box treatment effect into interpretable path-wise Shapley effects. We provide stand-alone theoretical results for decomposing the original on-manifold Shapley value (i.e. Shapley with a conditional reference distribution) of a treatment feature into path-wise local causal effects. We claim that our method achieves the four goals presented above: reliability, safety, interpretability and high resolution. Our contributions are as follows:
\setlist{nolistsep}
\begin{itemize}[leftmargin=4mm, itemsep=2mm,topsep=0pt]
\item We introduce Path-Wise Shapley (PWSHAP) effects, an extension of on-manifold Shapley values for locally explaining treatment effect under a causal DAG. Robustness to adversarial attacks (and thus safety) is guaranteed by the adoption of a conditional reference distribution. Reliability is ensured by the acknowledgment of the causal structure. As such, PWSHAP reconciles both safety and reliability. 
\item We show how our method can be used as a non-parametric alternative to mediation and bias analysis.
We further show how PWSHAP can be used for fairness studies when the causal graph involves a mixture of fair/unfair paths, and under randomised treatment to assess effect modification (also referred to as moderation). We further show that Causal Shapley \citeMain{heskes2020causal}, the closest method to ours, does not acknowledge moderation. 
\item We establish error bounds (i) from the outcome model to the Shapley values and PWSHAP effects (ii) from the Shapley values and treatment model (referred to as propensity score) to the PWSHAP effects. 
\end{itemize}
To the best of our knowledge, we are the first to interrogate the link between the Shapley feature importance of a treatment, and the standard notion of the treatment effect as defined within the causal inference literature, as the expected difference between potential outcomes under the two treatments. 

\section{Shapley Values}
\label{sec:shapley_intro}
Shapley values are a local feature attribution method. They quantify the importances of the features  $\{1,\ldots,\numfeats\}$ of a complex machine learning model $\blackbox:\mathbb{R}^{\numfeats}\rightarrow\mathbb{R}^{\numout}$ at an instance $\localobs\in\mathbb{R}^{\numfeats}$, given only black-box access to the model. The local prediction $\blackbox(x)$ is formulated as a sum of individual feature contributions: $\blackbox(x)=\phi^{\blackbox}_{0}(x)+\sum_{i=1}^{M} \phi^{\blackbox}_{i}(x)$, where $\phi^{\blackbox}_{j}(x)$ is the contribution of feature $j$ to $\blackbox(x)$ and $\phi^{\blackbox}_{0}(x)=\mathbb{E} [f(X))]$ is the averaged prediction with the expectation over the observed data distribution. The Shapley value of a feature $j$ captures the change in model outcome comparing the prediction when the feature value $x_j$ is included to when it's removed from the input. This change is computed from the difference in value function $v$ when setting feature $j$ equal to the instance feature value $x_j$, averaged over all possible coalitions $\includedfeats$ of features excluding feature $j$. If a feature is included in the coalition its value is set to the observed instance value $x_j$. To model feature removal, the value function takes the expectation of the black-box algorithm at observation $\localobs$ over the non-included features $\droppedfeats$ using a reference distribution $\refdistvar{}{\includedfeats}$ such that $v_f(\feat{\includedfeats},x)=\expect_{\refdistvar{}{\includedfeats}}[\blackbox({\localobs_{\includedfeats}, \varimputed_{\droppedfeats}})] \label{eq:expvalue}$ for $\droppedfeats:=\{1,\ldots,\numfeats\} \backslash \includedfeats$ and the operation $({\localobs_{\includedfeats}, \localobs_{\droppedfeats}})$ denoting the concatenation of its two arguments.
Binomial weights ${|\includedfeats|!(\numfeats-|\includedfeats|-1)!}/(m-1)!$ take account of all possible orderings. The Shapley value of feature $j$ is thus:
\begin{align*}
    \phi_j^\blackbox(x) 
    &=   \sum_{i=0}^{\numfeats-1}   \frac{1}{\numfeats\binom{\numfeats-1}{i}} \sum_{\substack{\includedfeats \not\owns j \\|S|=i}} [v_f(S \cup \{j\}, x) - v_f(S, x)],
\end{align*}
 i.e. $\phi_j^\blackbox(x) = \mathbb{E}_{p(S \mid j \notin S)} [\coalitionshapley{f}{S}{j}{x}]$ where  $\coalitionshapley{f}{S}{j}{x} := v_f(S \cup \{j\}, x) - v_f(S, x)$ and $ \forall j, \ p(S \mid j \notin S) = \nicefrac{1}{\numfeats\binom{\numfeats-1}{|S|}}$.
Shapley values have become a gold standard amongst explanation models due to their desirable properties (model agnostic, additive) and axioms (\textit{Symmetry}, \emph{Efficiency}, \emph{Linearity} and \emph{Dummy}---see Supplement \ref{subsec:shap_axioms} for details). However, the method has not been adopted in critical settings due to the considerable limitations of both possible reference distributions \citeMain{janzing2020feature, chen2020true, sundararajan2020many}.

\textbf{Limitations of Shapley values} \quad On the one hand, \textit{on-manifold} Shapley values \citeMain{aas2021explaining} use a conditional reference distribution, conditioning on $\localobs_{\includedfeats}$ to better account for correlations between features $\refdistvar{}{\includedfeats}:=\dist({\varimputed\given\varimputed_{\includedfeats}=\localobs_{\includedfeats}})$. Sampling from a conditional distribution forces the model to be evaluated on plausible instances that lie on the data manifold. It thus improves the adversarial robustness and thus the safety of the method \citeMain{slack2020fooling}. However, on-manifold Shapley values have been shown to be unreliable as they can generate misleading explanations \citeMain{janzing2020feature, sundararajan2020many}. On the other hand, \textit{off-manifold} Shapley values use a marginal reference distribution, that is $\refdistvar{}{\includedfeats}:=\dist({\varimputed})$ \citeMain{lundberg2017unified}. The resulting explanations reveal the functional dependence better, also known as being ``true to the model'' \citeMain{chen2020true}. However, sampling from the marginal distribution breaks the dependence between features. Consequently, off-manifold Shapley values are sensitive to adversarial robustness and thus deemed unsafe \citeMain{slack2020fooling}. Note that adversarial robustness is key for fairness studies. If an unfair model undergoes an adversarial attack, it may ``counterbalance'' its potential prejudice on real-world data by forming predictions favourable to disadvantaged groups on implausible inputs. Since only the marginal distribution is used, the resulting Shapley value of a sensitive attribute might look fair even though the model would predict unfairly on real-world data \citeMain{slack2020fooling}. Ultimately, Shapley values can't provide both reliable and safe explanations, which may hinder their adoption in safety-critical settings (see further details on Shapley values in Supplements \ref{sec:shapley_def}). 

Shapley values also have limited interpretability. 
The attribution of a target feature $j$ is the result of model evaluations averaged over all coalitions excluding $j$. The goal of this procedure is to acknowledge all the correlations amongst features. However, if some features are assumed to be independent, this assumption fails. Averaging over coalitions with/without independent features may generate redundancies and unbalance the resulting attribution. 
Also, the \emph{interpretation} of on-manifold and off-manifold Shapley values is agnostic to the assumed \emph{causal structure}, if any.  When a specific treatment is of interest, causal interpretation of its Shapley values should be done in light of the relative roles of other features: confounder, moderator or mediator; see Supplement \ref{sec:ci} for a definition of these notions. Interpreting Shapley values causally would be a case of 
the ``Table 2 fallacy'' \citeMain{westreich2013table}, where all coefficients of a model are misleadingly interpreted as adjusted causal effects. Thus we claim that under a posited DAG, the Shapley value of a feature should be computed according to the assumed statistical dependencies, i.e.\ the \emph{edges} in the DAG, and interpreted in light of its causal links with other variables, i.e.\ the \emph{directions} of the arrows in the DAG. 

In PWSHAP, we use a conditional reference distribution to ensure the robustness to adversarial attacks and safety of our method. Meanwhile, we are able to generate feature attributions that are both reliable and interpretable, thanks to the tailored causal interpretations of the effects we compute.


\section{Path-Wise SHAP (PWSHAP)} 
\label{sec:pwshap}
The intuition behind the introduced method is two-fold. First, we decompose the Shapley value as a weighted sum of quantities that can be interpreted causally as treatment effects along coalitions. Second, by only considering relevant coalitions, we are able to deduce quantities that can be interpreted causally along paths. Since the treatment $T$ is of special interest, we separate it from the other variables, that we call covariates $C$, such that $X=(C,T)$.
\subsection{Problem Setup}
\label{subsec:problem_Setup}
Let $C$ denote covariates, $T$ a binary treatment and $Y$ an outcome of interest. We assume that
$Y = f^*(C,T) + \epsilon$, with $\mathbb{E}[\epsilon|C,T] = 0$. 
Our black-box $f$ is an arbitrary function of $X = (C,T)$ which aims at predicting $f^*$. We aim at explaining the specific effect of the treatment variable $T$ on the predictions made by the black-box $f$ for an individual with values $c$ of covariates $C$. To do so, we first decompose the Shapley value of $T$ into a weighted sum of ``Shapley effects''  which are inspired by conditional average treatment effects, commonly used in the causal literature. We refer to a coalition $S$ excluding treatment $T$ as a \textit{subset of covariates} and note the value function as $v_f(S \cup \{T\},c_{S},t)$ when it is taken over the coalition $S \cup \{T\}$ and $v_f(S,c_{S})$ when taken over $S$. Notations are summarised in Section \ref{sec:notations}, with a running example to illustrate them all in Supplement \ref{sec:running_exp}.

PWSHAP relies on two assumptions: (i) the treatment of interest is a causal ancestor of the outcome (no anti-causal learning) and (ii) the DAG is compatible with the observed data i.e.\ all conditional independence constraints implied by graphical d-separation relations hold in the data. The user-supplied DAG thus only encodes the conditional dependences and is not assumed to be identical to the underlying model behavior. The ``direction'' of the arrows in the DAG is only used for causally \emph{interpreting} the PWSHAP values.

\subsection{Decomposition into Shapley Effects}
\label{sec:shapley_decomposition}
First, we notice a connection between value functions $v_f$ of the black-box $f$ and conditional average treatment effects using coalition-wise Shapley values.

\begin{definition}[Coalition-wise Shapley effect] \label{def:coalition_shapley_values}
We define the coalition-wise Shapley effect\footnote{Note that our Shapley effects are orthogonal to those introduced by \citeMain{iooss2017shapleyeffects} for numerical models .} of $T$ on $Y$ along the covariates $C_S$ indexed by the subset of covariates $S$ as:
\begin{align*}
    &\effect^f_{ T \rightarrow Y | C_{S}}(c_S) = v_{f}(S \mkern-3mu \cup \mkern-3mu \{T\},c_S,1) - v_{f}(S \mkern-3mu \cup \mkern-3mu \{T\},c_S,0) &&
\end{align*}
\end{definition}
 The coalition-wise Shapley effect can be understood as a generalisation of conditional average treatment effects. Indeed, for the true model $f^*$, the RHS is equal to $\mathbb{E}[Y|C_S = c_S, T=1] - \mathbb{E}[Y|C_S = c_S, T=0]$.
Under the typical causal treatment effect identification assumptions, i.e.\  no interference, consistency, and conditional exchangeability given $C$  \citeMain{imbens2010rubin}, this is the conditional average treatment effect (CATE) \citeMain{rubin2005causal} (definition in Supplement \ref{sec:ci}) when $S$ is the complete coalition, i.e.\ containing all covariates. In addition, $\effect^f_{ T \rightarrow Y |{\emptyset}}$ is the ``base'' treatment effect, i.e.\ a population-wide estimate of treatment effect. Its exact causal interpretation depends on the structure of the DAG, but in some cases it equates to the Average Treatment Effect (ATE) as defined by \citeMain{rubin2005causal} (definition in Supplement \ref{sec:ci}).
The coalition-specific Shapley effect can be linked to the original Shapley values as follows.
\begin{property}[Decomposing Shapley values into Shapley effects]  \label{prop:shapley_decomposition}
The \textit{coalition-wise Shapley value} $\coalitionshapley{f}{S}{T}{c,t}$ is equal to a weighted estimate of a local treatment effect, 
\begin{align}
    \coalitionshapley{f}{S}{T}{c,t} &= w^*_{S}(c_S,t) \cdot \effect^f_{ T \rightarrow Y | C_{S}}(c_S), 
\end{align}
where $w^*_{S}(c,t)$ denotes what we call the ``propensity weights'' defined by 
$w^*_{S}(c,t) = t - p(T=1|C_S=c_S)$. This name follows the fact that these weights are related to whether the sample is an outlier or not.

The proof can be found in Supplement \ref{proof:shapley_decomposition}.
Property \ref{prop:shapley_decomposition} shows that each coalition-specific term in the original on-manifold Shapley value is equal to the product of two terms. The first is a weight that depends on the propensity score. The second is a measure of the treatment effect, namely the coalition-specific Shapley effect. As a result, the overall Shapley value $\phi_T^f(c,t)$ can be decomposed as 
\begin{align*}
    \phi_T^f(c,t) &= \mathbb{E}_{p(S \mid T \notin S)} [w^*_{S}(c_S,t) \cdot \effect^f_{ T \rightarrow Y | C_{S}}(c_S) ].
\end{align*}
\end{property}

\subsection{Path-Wise Shapley (PWSHAP) Effects}
 Although we connected Shapley values to coalition-wise Shapley effects the latter still only apply to \textit{coalitions} and not specific \textit{paths}. However, the coalition-wise Shapley effect $\effect^f_{ T \rightarrow Y | C_{S}}(c_S)$ can be understood as the causal flow from $T$ to $Y$ through a set of covariates $S$. Thereby, we define the causal flow along the (undirected) path from $T$ to $Y$ through $C_i$ as the difference between the causal flow through all covariates and the causal flow through all covariates but $C_i$. See Supplement \ref{sec:generalisation} for a generalisation of paths of length 3 or more. 
\begin{definition}[Path-wise Shapley effect] \label{def:path_shapley_values}
Let $S^*$ be the coalition with all covariates. We refer to the following quantity as the path-wise Shapley effect of $T$ on $Y$ along the path from $T$ to $Y$ through $C_i$:
\begin{align*}
        &\effect^f_{C_i}(c) = \effect^f_{ T \rightarrow Y | C_{S^*}}(c) - \effect^f_{ T \rightarrow Y | C_{S^*\backslash \{i\} }}(c_{S^*\backslash \{i\}}) .
\end{align*}
\end{definition}
For instance in the fairness example from Section \ref{sec:intro}, the path-wise Shapley effect of sex on admission (Adm) mediated by the chosen department (Dpt) $\effect^f_{Sex \rightarrow Dpt \rightarrow Adm}$ is 
 $\effect^f_{Sex \rightarrow Adm| {Dpt, Exam}}  - \effect^f_{Sex \rightarrow Adm| {Exam}}$.

Path-wise Shapley effects thus quantify the change in model outcome when specifying the feature values along a specific path, compared to when all features are specified but the ones on the path of interest. As such, PWSHAP measures the effect of the treatment on the outcome through a causal pathway. Ultimately, conditioning on all other features reinforces the locality of our result. 
 It can also be seen as a contribution to the shift from a global estimated ``base'' treatment effect to an individual estimated treatment effect.
 However, note that PWSHAP violates the efficiency property i.e.\ they do not sum up to an interpretable quantity like the original Shapley feature attributions do. 
Moreover, Property \ref{prop:integration_pathwise_effect} shows that integrating PWSHAP effects can help isolate covariates that are conditionally independent on the treatment given other covariates (the Supplement \ref{proof:integration_pathwise_effect} for the proof). As shown in Section \ref{sec:causal_interpretation}, the actual causal meaning of this conditional independence depends on the posited DAG of the data, however.
\begin{property}[Integration of the PWSHAP effects] \label{prop:integration_pathwise_effect} 
Let $C_i$ be a covariate such that $C_i \indep T | C_{-i}$ where  $C_{-i} := C_{S^* \backslash \{i\}}$. Then for any function $f$ and any value $c_{-i}$ of $C_{-i}$, 
\begin{align*}
\mathbb{E}[\effect^f_{C_i}(C_i, c_{-i}) | C_{-i} = c_{-i} ] = 0
\end{align*}
\end{property}
\vspace*{-0.5cm}
\subsection{Estimation of Shapley Effects from Shapley Values}
\label{sec:estimation_procedure}
Using Property \ref{prop:shapley_decomposition}, we can express the coalition-wise Shapley effects $\effect^f_{ T \rightarrow Y | C_{S}}$ from the coalition-wise Shapley values $\coalitionshapley{f}{S}{T}{c,t}$ as
$\effect^f_{ T \rightarrow Y | C_{S}}(c_S) = \nicefrac{\coalitionshapley{f}{S}{T}{c,t}}{w^*_S(c,t)}$.
Therefore, the path-wise Shapley effects $\effect^f_{C_i}$ are computed as:
\begin{align*}
&\effect^f_{C_i}(c) = \frac{\coalitionshapley{f}{S^*}{T}{c,t}}{w^*_{S^*}(c,t)} - \frac{\coalitionshapley{f}{S^* \backslash \{i\}}{T}{c,t}}{w^*_{S^* \backslash \{i\}}(c,t)}.
\end{align*}
In practice, path-wise Shapley effects are computed by replacing the true propensity weights with weights that use an estimate of the propensity score. For this, we further need to assume positivity holds. The path-wise Shapley effect of $T$ on $Y$ through $C_i$ is thus estimated in three steps: (i) computing the coalition-wise Shapley values for $S^*$ the entire set of covariates and $S^* \backslash \{i\}$; (ii) dividing each of these terms by an estimate of their corresponding propensity weight; (iii) taking the difference between the two resulting quantities (also known as coalition-specific Shapley effects) to isolate the effect along the path through $C_i$. Note that division by weights requires overlap, that is $\forall c, 0 < p(T=1|C=c) < 1$.



\section{Related Work}
\subsection{Conceptual Distinction Between Local Explanations Models and Causal Inference}
Causal inference aims at assessing the effect of a \emph{feature} at a global scale (e.g. ATE) or within subgroups (e.g. CATE). Contrastingly, Shapley values assess the \emph{local} effect of a \emph{feature value} compared to the values taken by that feature in the reference distribution. Therefore, to identify path-wise local effects, we consider a path to be ``deactivated'' when the covariate value gets sampled from the reference distribution, i.e.\ the covariate is \emph{not} in the coalition. Conversely, specifying a covariate value, i.e.\ when the covariate \emph{is} in the coalition, ``activates'' a path. Thereby, coalition-wise effects are conditional treatment effects marginalised over covariates that aren't in the coalition. In the admission example from Section \ref{sec:intro}, the coalition-specific Shapley effect for $\{Exam\}$, $\effect^f_{Sex \rightarrow Adm| \{Exam\}}$ corresponds to the treatment effect along two paths: the direct path $Sex \rightarrow Adm$ and the path from $Sex$ to $Adm$ through $Exam$. 
\subsection{Comparison of PWSHAP with Existing Methods}
\label{sec:comparison}
We compare our method to two baseline explanation methods. Our first baseline is Causal Shapley (CS) \citeMain{heskes2020causal}, another method aiming to explain a model under an assumed causal DAG. Like PWSHAP, Causal Shapley splits Shapley attributions, although the split is binary (direct/indirect effect). In Causal Shapley, the indirect effect of a feature $j$, the distribution of the ‘out-of-coalition’ features changes due to the do-operator (see Suppl. \ref{sec:shapley_def} and \ref{subsec:causal_shap} for further details). Our second baseline is on-manifold Shapley, a natural choice given that PWSHAP augments the original method. Section \ref{subsec:further_graph_based} details other graph based Shapley methods \citeMain{wang2021shapley, singal2021flow}, which are not appropriate baselines here due to structural differences. 

\textbf{Higher model fidelity, lower reliance on causal assumptions than Causal Shapley} We claim that PWSHAP has higher model fidelity and relies less on the assumed causal structure than Causal Shapley. As the direct/indirect effect split is based on \textit{do}-calculus in Causal Shapley, the computation of the attributions depends on the assumed DAG (both the edges and their directions). In contrast, PWSHAP computations only depend on the hypothesised feature \emph{dependencies} i.e.\ the edges in the DAG. Only the causal \emph{interpretation} of PWSHAP depends on the direction of the edges. We view the fact that our approach is agnostic to the choice of a (compatible) DAG as a strength, as it allows different experts to explain the black-box model output according to their own causal beliefs about the data or phenomenon being studied (see \ref{subsec:further_dag} for a detailed discussion on this). Ultimately, by applying \textit{do}-calculus, Causal Shapley computes feature attributions according to preconceptions of how the model should reason, and as such is “forcing” explanations to fit to a presumed causal structure. 
To further illustrate the limitations of relying on the causal assumptions and show that PWSHAP has higher fidelity to the model, let us consider a black-box with a single covariate $C$, and a treatment $T$. If we wrongly assume $C$ to be a confounder instead of a mediator, the indirect effect of treatment i.e.\ the mediation of treatment through $C$ would be null according to Causal Shapley (see Property \ref{prop:indirect_part} in Supplement \ref{subsec:causal_shap}). By contrast, only the causal interpretation of the PWSHAP effect through $C$ would be incorrect, but its value would remain unaltered. 

\textbf{Increased resolution, better interpretability}
Compared to both Causal and on-manifold Shapley, PWSHAP has higher resolution---as it is path-specific instead of feature-specific---and improved interpretability. In Causal Shapley and on-manifold Shapley values, feature attributions result from taking an average over coalitions, whereas PWSHAP only considers coalitions used to compute effects. Ultimately, evidence has shown that on-manifold Shapley values and Causal Shapley values can generate misleading interpretations \citeMain{sundararajan2020many}. In on-manifold Shapley values the attribution of a feature that does not appear in the algebraic formulation of the model can be non-zero, depending on how the data is distributed. This is induced by both the conditional reference distribution, the average taken over multiple coalitions. By providing an exact interpretation for the computed quantities, PWSHAP overcomes this unreliability issue. If a PWSHAP effect $\effect^f_{C_i}$ is null, it means that specifying the covariate $C_i=c_i$ has had no impact on the treatment effect compared to marginalising it, \emph{according to our black-box} (see Lemma \ref{lemma:local_confounding_effect} and
Property \ref{prop:local_mediator_effect_confounders}). Meanwhile, PWSHAP remains robust to adversarial attacks, as it samples from a conditional reference distribution \citeMain{slack2020fooling}. PWSHAP thus reconciles \textit{safety} and \textit{reliability}
. However, a limitation of PWSHAP compared to both baselines is that it violates the efficiency property (see Suppl. \ref{sec:further}).

\section{Error Bounds} 

\label{sec:error_bounds}
\begin{figure}
\centering
{
\scalebox{0.9}{
\begin{tikzpicture}[node distance=15mm, >=stealth]
 \pgfsetarrows{latex-latex};
\begin{scope}[xshift=2.5cm]
\node[rve] (T) {$T$};
\node[above of=T, yshift=-4mm] (number) {$(2)$};
\node[rve, above right of=T, yshift=-4mm] (C1) {$C_1$};
\node[rve, below right of=T, yshift=4mm] (C2) {$C_2$};
\node[rve, above right of=C2, yshift=-4mm] (Y) {$Y$};
 \draw[deg] (C1) to (T);
 \draw[deg] (C2) to (T);
 \draw[deg] (C1) to (Y);
 \draw[deg] (T) to (Y);
 \draw[deg] (C2) to (Y);
\end{scope}
\begin{scope}
\node[rve] (T) {$T$};
\node[above of=T, yshift=-4mm] (number) {$(1)$};
\node[rve, right of=T] (Y) {$Y$};
\node[rve, yshift=8mm] (A) at ($(T)!.45!(Y)$) {$C_1$};
\node[rve, yshift=-8mm] (B) at ($(T)!.45!(Y)$) {$C_2$};
\node[inner sep=0] (m) at ($(T)!.45!(Y)$) {};
\draw[deg] (T) to (Y);
\draw[deg] (A) to (m);
\draw[deg] (B) to (m);
\end{scope}
 \begin{scope}[xshift=5.5cm]
 \node[rve] (T) {$T$};
 \node[rve, above right of=T, yshift=-4mm] (C1) {$C_1$};
 \node[rve, below right of=T, yshift=4mm] (C2) {$C_2$};
 \node[rve, above right of=C2, yshift=-4mm] (Y) {$Y$};
  \node[above of=T, yshift=-4mm] (number) {$(3)$};
 \draw[deg] (T) to (C1);
  \draw[deg] (T) to (C2);
  \draw[deg] (C1) to (Y);
  \draw[deg] (T) to (Y);
  \draw[deg] (C2) to (Y);
  \end{scope}
\end{tikzpicture}
}
 \label{fig:dag_blocks}
}
\newline
\newline
{
\includegraphics[width=8cm]{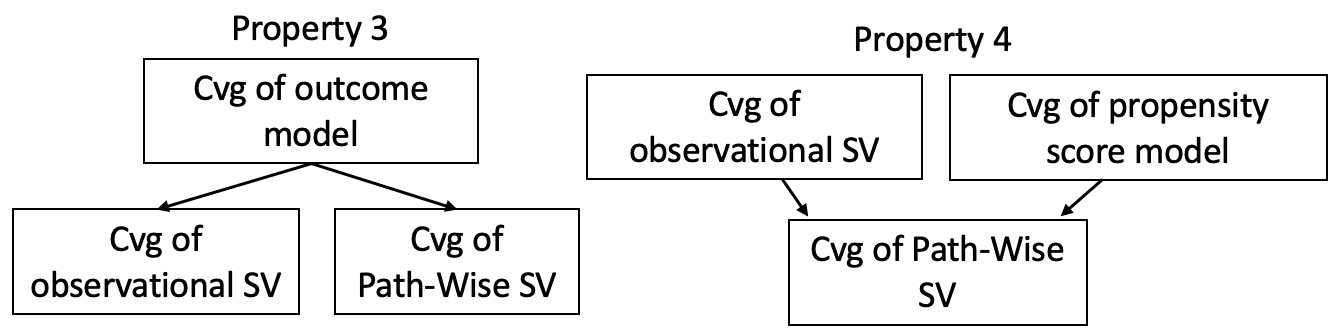}
\label{fig:infography}
}
\caption{DAGs for Building Blocks (Up) and Error Bound  (Down, Cvg=Convergence, SV=Shapley value)}
\label{fig:dag_blocks_infography}
\end{figure}
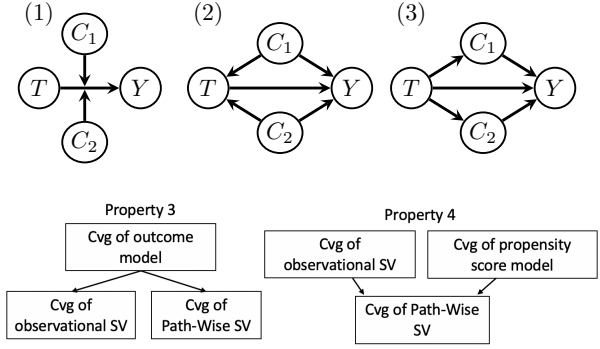

We now show how to obtain error bounds for quantities like path-wise Shapley effects from other quantities like the outcome model, according to Figure \ref{fig:dag_blocks_infography}. To the best of our knowledge, these are the first results regarding error bounds for on-manifold Shapley values.
In the following, $(\hat{f}_N)$ denotes a sequence of estimators of $f^*$. The proofs can be found in Supplements \ref{proof:error_bounds_outcome} and \ref{proof:error_bounds_shapley}

\begin{property}[Convergence of the outcome model implies convergence of Shapley values and PWSHAP effects] \label{prop:error_bounds_outcome} 
If $\forall c, t, N, | \hat{f}_N(c,t) - f^*(c,t) | \leq e^\text{outcome}_N$ then:
\begin{enumerate}[leftmargin=4mm]
    \item Convergence of the coalition-specific Shapley terms: $\forall c, t, N, \qquad | \coalitionshapley{\hat{f}_N}{S}{T}{c,t} -  \coalitionshapley{f^*}{S}{T}{c,t} | \leq 2 e^\text{outcome}_N.$
    which implies the convergence of the Shapley value of the estimated model to that of the true model.
    \item Convergence of the path-wise Shapley effects: $\forall i, c, N, \qquad | \effect^{\hat{f}_N}_{C_i}(c) - \effect^{f^*}_{C_i}(c) | \leq 4 e^\text{outcome}_N.$
\end{enumerate}
\end{property}

\begin{property}[Convergence of estimated coalition-specific Shapley values and propensity score implies convergence of estimated PWSHAP effects] \label{prop:error_bounds_shapley} 
Assuming that: \\
(1) the arbitrary propensity score model $\pi^N$ and the true propensity score model $\pi^*$ verify $\epsilon$-strong overlap, \\
(2) $\forall c, N |\pi^N(c) - \pi^*(c) | \leq e^\text{propensity}_N$ \\
(3) $\forall S \text{ s.t. } T \notin S, c, N,  | \estcoalitionshapley{N}{\hat{f}_N}{S}{T}{c,t} - \coalitionshapley{f^*}{S}{T}{c,t} | \leq e^\text{Shap}_N$,

with $w^N_S(c,t) = t - \mathbb{E}_p(C_{\bar{S}}|C_S=c_S)[\pi^N(c_S,C_{\bar{S}})]$ we show the convergence of the estimated PWSHAP effects to the true PWSHAP effects, 
     $\forall i, c, t, N,$ 
    \begin{align*}
           | \hat{\effect}^{N,\hat{f}^N}_{C_i}(c) - \effect^{f^*}_{C_i}(c) | \leq \frac{4 e^\text{Shap}_N}{\epsilon} + \frac{4||f^*||_\infty \cdot e^\text{propensity}_N}{\epsilon^2} .
    \end{align*}

\end{property}

\section{Causal Interpretations}
\label{sec:causal_interpretation}
PWSHAP effects are interpreted by revisiting causal inference concepts of confounding, moderation and mediation at a local scale. As our method stands on theoretical grounding, we first provide objective evidence using explicit equations. 



\subsection{Local Bias Analysis} \label{sec:local_bias}
Under DAG (2) of Figure \ref{fig:dag_blocks_infography}, PWSHAP effects are causally interpreted as follows:
\begin{align*}
\phi_T^{f^*} &= \nicefrac{w^*_{12}}{3} \cdot \text{CATE}(c_1,c_2) + \nicefrac{w^*_1}{6} \cdot \text{ "CATE"}_{C_1}(c_1) \\
&+ \nicefrac{w^*_2}{6} \cdot \text{"CATE"}_{C_2}(c_2) + \nicefrac{w^*}{3} \cdot \text{Diff. in means}
\end{align*}
where $\text{"CATE"}_{C_S}(c_s) = \mathbb{E}[Y|T=1,C_S=c_S] - \mathbb{E}[Y|T=0,C_S=c_S]$.  We refer to these terms as ``CATE''s, in an abuse of notation, but note that they are not true causal conditional average treatment effects, as they include confounded paths. The term ``Diff. in means'' stands for  $E[Y|T=1]-E[Y|T=0]$. To isolate the spurious effect of the confounders, we further assume that the two confounders are not effect moderators.
\begin{definition}[Local confounding effect] \label{def:local_confounding_effect} In this example, we call the PWSHAP effect of $C_2$ the ``local confounding effect of $C_2$'' and note it $\effect^f_{T \leftarrow C_2 \rightarrow Y}$. In other words, $\effect^f_{T \leftarrow C_2 \rightarrow Y} := \effect^f_{C_2}$
\end{definition}
Notably, for the true model $f^*$, $\effect^{f^*}_{T \leftarrow C_2 \rightarrow Y}(c_1,c_2) = \text{CATE}(c_1,c_2) - \text{"CATE"}_{C_1}(c_1)$. This quantity has been referred to as the bias due to unmeasured confounding (assuming we observe $C_1$ but not $C_2$) in a segment of the sensitivity analysis literature \citeMain{veitch2020sense}. Therefore, our measure of confounding effect is a local equivalent of this bias. Indeed, integrating out this difference over $C_1,C_2$ yields $\text{ATE} - \mathbb{E}[\mathbb{E}[Y|T=1,C_1] - \mathbb{E}[Y|T=0,C_1]]$
where ATE is the Average Treatment Effect. However, if a covariate is both a confounder and an effect modifier, its path-wise attribution will cover both phenomena and the two effects will be indiscernible. Ultimately, PWSHAP contrasts with sensitivity analysis methods which are meant for quantifying unobserved confounding, whereas our method measures the impact of an observed confounder. 
 Further details on such techniques can be found in the Supplement \ref{sec:other_analyses}.
\begin{lemma}[Integration of the local confounding effect, true model] \label{lemma:local_confounding_effect}
 Let $C_1, C_2$ be two pre-treatment covariates such that ignorability given $C_1, C_2$ holds, i.e.\ $\forall t, Y(t) \indep T | C_1,C_2$. If, additionally, $C_2$ is not a confounder, i.e.\ $C_1$ alone guarantees ignorability or $\forall t, Y(t) \indep T | C_1$, then the integral of the local confounding effect of $f^*$ w.r.t. $C_2$ on the joint distribution of covariates is null: \\
\centerline{$\mathbb{E}[\effect^{f^*}_{T \leftarrow C_2 \rightarrow Y}(C_1,C_2)] = 0$.}
\end{lemma}

The proof can be found in Supplement \ref{proof:local_confounding_effect}. For a variable that is not actually a confounder, the integration of the local confounding effect thus yields zero. This can be generalised to any number of confounding pre-treatment covariates, by grouping all of them in $C_1$. For any blackbox $f$, a stricter condition yields the same result as a corollary of Proposition \ref{prop:integration_pathwise_effect}. We give that result and an example of local bias analysis in Supplement \ref{subsec:local_bias}. Further, if the local confounding effect is zero for all individuals in the training set, then we can hypothesise that the model did not learn to predict through the confounding path $T \leftarrow C_2 \rightarrow Y$. 

\subsection{Local Moderation Analysis Under Randomised Treatment}
\label{sec:local_moderation}
Here, ``moderation'' refers to an effect modification as in \citeMain{boruvka2018assessing}. 
In the setting represented in Figure \ref{fig:dag_blocks_infography}, causal graph (1), where treatment is assumed to be unconfounded, we interpret the PWSHAP decomposition as follows:
\begin{align*} 
    \phi_T^{f^*}
    &= \nicefrac{1}{3} \cdot w^*_{12} \cdot \text{CATE}(c_1,c_2) + \nicefrac{1}{6} \cdot w^*_1 \cdot \text{CATE}_{C_1}(c_1) \\
    &\ \ \ + \nicefrac{1}{6} \cdot w^*_2 \cdot \text{CATE}_{C_2}(c_2) + \nicefrac{1}{3} \cdot w^* \cdot \text{ATE}
\end{align*} 
\begin{definition}[Local moderating effect] \label{def:local_moderating_effect} In this example, we call the PWSHAP effect of $C_2$ the ``local moderating effect of $C_2$'' and denote it $\effect^f_{C_2: T \rightarrow Y}$. In other words, $
\effect^f_{C_2: T \rightarrow Y} := \effect^f_{C_2}$.
\end{definition}
PWSHAP assesses the local effect modification induced by $C_2$ by ``unspecifying'' this feature. 
Having null local moderating effect would mean that $C_2$ did not act as a moderator for this specific subject, \emph{according to our fitted black-box}. 
Unlike previous methods \citeMain{imai2013estimating, athey2016recursive, wang2017causal}, our PWSHAP approach to moderation analysis does not involve subgroup finding---a technique known to be under-powered \citeMain{holmes2018machine}---and is nonparametric (see Supplement \ref{sec:other_analyses} for a review of moderation analysis). Ultimately, we show that in the presence of pre-treatment moderators, Causal Shapley compounds the main effect of treatment and its effect via moderation into a single ``direct'' effect, whereas PWSHAP explanations are able to distinguish the added treatment effect due to moderation from the main effect. 
We compare PWSHAP with Causal Shapley on an example as shown in DAG (1) of Figure \ref{fig:dag_blocks_infography} assuming $Y = \beta T + \gamma_1C_1 + \gamma_2C_2 + \alpha_1TC_1 + \alpha_2TC_2 + \epsilon$ with $\mathbb{E}[\epsilon|T,C_1,C_2] = 0$ and where $C_1$, $C_2$ are two independent moderators with $C_1, C_2 \sim \text{Uniform}(0,1)$. Treatment is randomised: $T \sim \text{Bernoulli}(p)$. Details about the following are given in Supplement \ref{subsec:local_moderation}. PWSHAP yields:
\begin{align*}
&\effect^{f^*}_{T \rightarrow Y | C_1, C_2} = \beta + \alpha_1c_1 + \alpha_2c_2 \\
&\effect^{f^*}_{C_1} := \effect^{f^*}_{C_1: T \rightarrow Y} = \alpha_1 (c_1 - \nicefrac{1}{2}) \\
&\effect^{f^*}_{T \rightarrow Y | \emptyset} = \beta + \nicefrac{\alpha_1}{2} + \nicefrac{\alpha_2}{2} \\
&\effect^{f^*}_{C_2} := \effect^{f^*}_{C_2: T \rightarrow Y} = \alpha_2 (c_2 - \nicefrac{1}{2}) 
\end{align*}
where $\expect [C_1] = \expect [C_2] = \nicefrac{1}{2}$. PWSHAP effects through $C_1$ and $C_2$ are null if $C_1=C_2= \nicefrac{1}{2}$. The PWSHAP approach thus matches the default behaviour of local explanation methods: paths through effect moderators are given zero attribution if the moderator value is equal to the population average. This highlights the true locality of our method. Furthermore, one can check that the moderating effects integrate to $0$. Again, this is coherent with the overall definition of randomised treatment in causal inference. By contrast, moderation by $C_1$ and $C_2$ is overlooked in Causal Shapley as $\phi^{f^*, \text{CS}}_{T, \text{indirect}} = 0$. Further, $\phi^{f^*, \text{CS}}_{T, \text{direct}} = (t - p)\{\beta + \frac{\alpha_1}{2} \cdot (c_1 + \frac{1}{2}) + \frac{\alpha_2}{2} \cdot (c_2 + \frac{1}{2})\}$ which does not reflect the local behaviour of the model.
\subsection{Local Mediation Analysis}
\label{sec:local_med}
Under DAG (3) of Figure \ref{fig:dag_blocks_infography}, i.e.\ with unconfounded treatment and two mediators only depending on it, the causal interpretation of the PWSHAP approach to mediation is:
\begin{align*}
\phi_T^{f^*} &= \nicefrac{1}{3} \cdot w^*_{12} \text{CDE}_{C_1,C_2}(c_1,c_2) + \nicefrac{1}{6} \cdot w^*_2 \text{CDE}_{C_2}(c_2)\\
&+ \nicefrac{1}{6} \cdot w^*_1 \text{CDE}_{C_1}(c_1) + \nicefrac{1}{3} \cdot w^* \text{ATE}
\end{align*}
where CDE refers to the \textit{Controlled Direct Effect} (definition in Supplement \ref{sec:ci}), with $\text{CDE}_{C_S}(c_s) = \mathbb{E}[Y|T=1,C_S=c_s] - \mathbb{E}[Y|T=0,C_S=c_s]$. 
We claim that the difference in CDE is able to isolate the local effect of a given mediator and has a causal interpretation, as outlined by the local mediating effect introduced below. 


\begin{definition}[Local mediating effect] \label{def:local_mediating_effect}
Here, we call the PWSHAP effect of $C_2$ the ``local mediating effect of $C_2$'' and note it $\effect^f_{T \leftarrow C_2 \rightarrow Y}$. So, $\effect^f_{T \rightarrow C_2 \rightarrow Y}  := \effect^f_{C_2}$.
\end{definition}

\begin{property}[Ancestors of outcome] \label{prop:local_mediator_effect_confounders} Let $M_1, M_2$ be two post-treatment and pre-outcome variables. 
Assuming that variables $C$ include all confounders of the relationships between $T$, $Y$ and $(M_1,M_2)$ and that $M_2 \indep T, M_1 | C$, then for any value $c$ of $C$ and $m_1$ of $M_1$,
\begin{align*}
\mathbb{E}[\effect_{T \rightarrow M_2 \rightarrow Y}(c,m_1,M_2) \mid C=c] = 0.
\end{align*}
\end{property}
In other words, if $M_2$ is not mediating the effect of $T$ on $Y$ because $M_2 \indep T | C$, and $M_2$ is independent of $M_1$ conditionally on $T, C$, then integrating the local mediating effect of $M_2$ yields 0 which is coherent with our intuition. The proof can be found in Suppl. \ref{proof:local_mediator_effect_confounders}. 
See Suppl. \ref{sec:med_analysis} for further comparisons with the traditional Natural Effects approach (definition in Supplement \ref{sec:ci}) and with Causal Shapley.

\section{Experiments: Synthetic Data}
\label{sec:exp_synth}

In the following two experiments, we show PWSHAP's ability to capture confounding and mediation on synthetic datasets. We infer path-specific Shapley effects following the procedure from Section \ref{sec:estimation_procedure}  and compute the absolute values of their averages $\bar{\effect}$ across the testing set. We divide these values by the empirical standard deviation of outcome $\sigma_Y$ on the training set to mitigate variation due to the scale of the outcome. We follow the same process for Causal Shapley's direct and indirect effects w.r.t.~the treatment. Results are averaged over 25 randomly sampled datasets, with standard errors shown in parentheses. More details are given in Supplement \ref{sec:exp_details}.  

\paragraph{Local bias analysis.} We consider the previous model with two pre-treatment covariates $C_1$ and $C_2$ described in DAG (2) of Figure \ref{fig:dag_blocks_infography}, and with results derived in Section \ref{sec:local_bias}.  We look at three scenarios: (i) neither $C_1$ nor $C_2$ are confounders, (ii) $C_1$ is a confounder but $C_2$ is not, (iii) both are confounders. Results are shown in Table \ref{tab:synthetic_xp_localbias_twovars}. Local confounding effects are significantly higher for confounding variables compared to non-confounding variables. This shows how these effects can isolate individual confounders in pre-treatment covariates, in accordance with Lemma \ref{lemma:local_confounding_effect}. Conversely, we do not notice any significant change in Causal Shapley's direct effect. Causal Shapley's indirect effect is even lower - as it is expected to be zero from Property \ref{prop:indirect_part}.

\begin{table}
\centering
\setlength{\tabcolsep}{2pt}
  \caption{Results on local bias analysis. 
  }%
\label{tab:synthetic_xp_localbias_twovars}

\begin{tabular}{l|cccc}
Scenario &  $\frac{|\bar{\effect}^f_{T \leftarrow C_1 \rightarrow Y}|}{\sigma_Y}$ &  $\frac{|\bar{\effect}^f_{T \leftarrow C_2 \rightarrow Y}|}{\sigma_Y}$ & $\frac{|\bar{\phi}^{f,\text{direct, CS}}_T|}{\sigma_Y}$ & $\frac{|\bar{\phi}^{f,\text{indirect, CS}}_T|}{\sigma_Y}$ \\ 
\toprule
 \makecell[l]{$C_1, C_2$ \\ non-conf.}  & \makecell{0.057 \\ (0.011)} & \makecell{0.064 \\ (0.013)} & \makecell{0.069 \\ (0.010)} & \makecell{0.006 \\ (0.001)} \\ 
 \midrule
 \makecell[l]{$C_1$ conf.,\\$C_2$ not} & \makecell{0.505 \\ (0.057)}  & \makecell{0.054 \\ (0.009)} & \makecell{0.067 \\ (0.008)} & \makecell{0.002 \\ (0.000)} \\ 
 \midrule
 \makecell[l]{$C_1, C_2$ \\ conf.} & \makecell{0.322 \\ (0.037)} & \makecell{0.277 \\ (0.034)} & \makecell{0.076 \\ (0.010)} & \makecell{0.006 \\ (0.003)}
\end{tabular}
\end{table}

\paragraph{Local mediation analysis.} We consider DAG (3) of Figure \ref{fig:dag_blocks_infography}, applied to a college admission example where we investigate the effect of sex---noted $T$ --- on the \textit{logit} of the probability of admission, mediated by exam results $C_1 = Q$ and department choice $C_2 = D$. Details are in Supplement \ref{subsec:local_mediation}.
We look at three scenarios : (i) neither is a mediator, (ii)  the former is a mediator but the latter is not, (iii) both are mediators. Results are presented in Table \ref{tab:synthetic_xp_localmediation_twovars}. Local mediating effects are significantly higher for mediating variables compared to non-mediating variables. This shows that these effects can isolate individual mediators in post-treatment covariates, in accordance with Property \ref{prop:local_mediator_effect_confounders}. Conversely, Causal Shapley's indirect effect seems to capture the presence of mediators - but not which variables are mediators.

\begin{table}
\centering
  \caption{Results on local mediation analysis. 
  }%
\label{tab:synthetic_xp_localmediation_twovars}
\setlength{\tabcolsep}{2pt}
\begin{tabular}{l|cccc}
Scenario &  $\frac{|\bar{\effect}^f_{T \rightarrow Q \rightarrow Y}|}{\sigma_Y}$ & $\frac{|\bar{\effect}^f_{T \rightarrow D \rightarrow Y}|}{\sigma_Y}$ & $\frac{|\bar{\phi}^{f,\text{direct, CS}}_T|}{\sigma_Y}$ & $\frac{|\bar{\phi}^{f,\text{indirect, CS}}_T|}{\sigma_Y}$ \\ 
\toprule
\makecell[l]{$Q, D$ \\ non-med.} & \makecell{0.056 \\ (0.012)} & \makecell{0.041 \\ (0.007)} & \makecell{0.072 \\ (0.012)} & \makecell{0.033 \\ (0.003)} \\ 
\midrule
\makecell[l]{$D$ med.,\\$Q$ not} & \makecell{0.059 \\ (0.010)} & \makecell{0.715 \\ (0.073)} & \makecell{0.080 \\ (0.011)}  & \makecell{0.091 \\ (0.019)} \\ 
\midrule
\makecell[l]{$Q, D$ \\ med.} & \makecell{0.948 \\ (0.117)} & \makecell{0.354 \\ (0.065)} & \makecell{0.112 \\ (0.014)} & \makecell{0.089 \\ (0.015)}   \\
\end{tabular}
\vspace*{-0.5cm}
\end{table}

\subsection{Robustness to Adversarial Attacks}

\begin{figure*}[t]
    \centering
    \begin{minipage}{0.24\textwidth}
      \centering
      (a)
      \includegraphics[width=\linewidth, height=3.2cm]{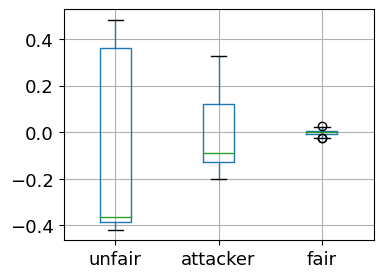}
    \end{minipage}%
    \begin{minipage}{0.24\textwidth}
      \centering
      (b)
      \includegraphics[width=\linewidth, height=3.2cm]{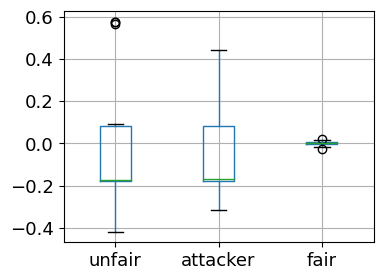}
    \end{minipage}%
    \begin{minipage}{0.24\textwidth}
      \centering
      (c)
      \includegraphics[width=\linewidth, height=3.2cm]{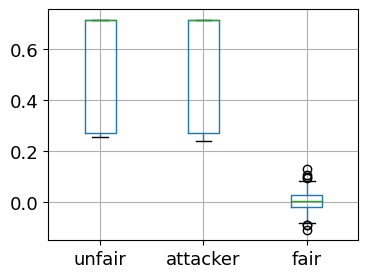}
    \end{minipage}
    \begin{minipage}{0.24\textwidth}
      \centering
      (d)
      \includegraphics[width=\linewidth, height=2.9cm]{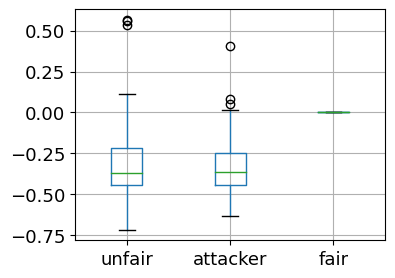}
    \end{minipage}
    \caption{Boxplots of (a) off-manifold Shapley values w.r.t.~$T$, (b) on-manifold Shapley values w.r.t.~$T$, (c) base PWSHAP effects $\effect_{ T \rightarrow Y |{\emptyset}}$,
    (d) PWSHAP effects on the path through $D$ $\effect_{ T \rightarrow D \rightarrow Y}$,
    each for the unfair, attacker and fair models.}
    \label{fig:adversarial}
\end{figure*}

We empirically show that on-manifold Shapley values are more robust to adversarial attacks than off-manifold Shapley values, as they are computed using a conditional reference distribution. Consequently, PWSHAP effects which marginalise over a conditional reference distribution are also more robust to adversarial attacks than off-manifold Shapley values. We compare the three explanation methods on a synthetic fairness study. Here, we consider three black-box models: a fair model, an unfair model, and an ``attacker model'' that returns fair predictions on instances classified as on the data manifold and unfair predictions on instances classified as not belonging to the data manifold. We generate a gender sensitive attribute as $T \sim \textit{U}(0,1)$, a departmental difficulty indicator $D \sim \textit{Bernoulli}(\pi (1-T) + (1-\pi)T )$, with $\pi = 0.99$. We also define a continuous test result $Q \sim \textit{U}(0,1)$. Notably, $D = 1-T$ with high probability, so we take a classifier classifying a given unit $(q,d,t)$ as belonging to the manifold iff $t = 1 - d$.  The fair model is defined $f^{\text{fair}}(t,q,d) = q$, the unfair model as $f^{\text{unfair}}(t,q,d) = \frac{t + td}{2}$, and the attacker model as $f^{\text{attacker}}(t,q,d) = 1_{\{t = 1 - d\}}f^{\text{unfair}}(t,q,d) + 1_{\{t = d\}}f^{\text{fair}}(t,q,d)$.
Figure \ref{fig:adversarial} shows the explanations given by off-manifold Shapley values w.r.t.~$T$, on-manifold Shapley values w.r.t.~$T$, base PWSHAP effects $\effect_{ T \rightarrow Y |{\emptyset}}$, and PWSHAP effets wrt the path $T \rightarrow D \rightarrow T$ $\effect_{ T \rightarrow D \rightarrow Y}$ on for the three types of black-box models in Figure \ref{fig:adversarial}. The on-manifold Shapley values and PWSHAP effects return very similar boxplots for the unfair and attacker models. Both methods also capture that the unfair model and the attacker model make predictions from the sensitive attribute $T$ while generating a low attribution to $T$ for the fair model. However, note that the boxplot of $\effect_{ T \rightarrow D \rightarrow Y}$ for the unfair model does not match the theoretical expectation (from Suppl. \ref{subsec:local_mediation}), a limitation that is most likely due to use of a potentially imprecise iterative imputer to fit conditional distributions, combined with division by small weights. Fitting conditional probabilities well to impute missing values remains an active topic of research \citeMain{lin2020missing}.
For the attacker model, the off-manifold Shapley values are between that for the unfair model and the fair model, as approximately half of the data with columns generated independently is classified as belonging to the manifold and half is not. This illustrates how off-manifold Shapley values are less robust to adversarial attacks than on-manifold Shapley values and derived methods like PWSHAP effects.

\section{Experiments : Real-World Data}
\label{sec:exp_uci}
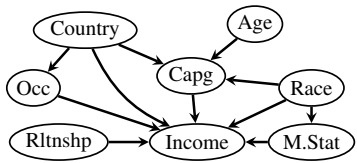
\begin{figure}
\centering
\scalebox{0.8}{
\begin{tikzpicture}[node distance=15mm, >=stealth]
 \pgfsetarrows{latex-latex};
\begin{scope}
\node[rve] (Y) {Income};
\node[rve, right of=Y, xshift=4mm, yshift=0mm] (M) {M.Stat};
\node[rve, above of=M, xshift=0mm, yshift=-6mm] (R) {Race};
\node[rve, left of=R, xshift=-5mm, yshift=2mm] (C) {Capg};
\node[rve, above left of=C, yshift=-3mm, xshift=-7mm] (Co) {Country};
\node[rve, above right of=C, yshift=-2mm] (A) {Age};
\node[rve, below left of=Co, yshift=1mm, xshift=2mm] (O) {Occ};
\node[rve, left of=Y, xshift=-8mm, yshift=0mm] (Re)  {Rltnshp};
\draw[deg] (R) to (Y);
\draw[deg] (R) to (M);
\draw[deg] (R) to (C);
\draw[deg] (M) to (Y);
\draw[deg] (C) to (Y);
\draw[deg] (Re) to (Y);
\draw[deg] (A) to (C);
\draw[deg] (O) to (Y);
\draw[deg] (Co) to[bend right=20] (Y.150);
\draw[deg] (Co) to (C);
\draw[deg] (Co) to (O);
\end{scope}
\end{tikzpicture}
}
\caption{DAG of the Census Income Dataset.}
\label{fig:dag_uci}
\end{figure}

We present a local mediation analysis experiment on the Adult data set from UCI \citeMain{asuncion2007uci}, using the causal graph from \citeMain{frye2019asymmetric}. Further results and experimental details can be found in the Supplement \ref{sec:exp_details}.
The binary outcome denotes whether an individual's income exceeds \$50,000 per year. 
The causal structure of the data is described in the DAG in Figure \ref{fig:dag_uci}. Race was dichotomised into white/non-white. Our individual of interest is a white 38-year-old born in the US, whose marital status (M.Stat) is divorced and Relationship (Rltnshp) is unmarried, and who has a managerial occupation and no capital gain (Capg).

\begin{table}
\centering
  \caption{Results on Census Income Data. 
  }%
\label{tab:results_uci}
\begin{tabular}{lll}
\hline
\footnotesize{Causal SHAP} & \footnotesize{$\phi_{\text{direct}}$}   & \footnotesize{$<0.001  (0.003)$ } \\
            & \footnotesize{$\phi_{\text{indirect}}$}   & \footnotesize{$0.004  (0.006)$} \\ \hline
\footnotesize{PWSHAP}      & \footnotesize{$\effect_{\text{Race} \xrightarrow{\text{total}} \text{Inc}}$}  & \footnotesize{$-0.005  (0.003)$} \\
            & \footnotesize{$\effect_{\text{Race} \rightarrow \text{Capg} \rightarrow \text{Inc}}$} & \footnotesize{$0.077  (0.004)$}   \\
            & \footnotesize{$\effect_{\text{Race} \rightarrow \text{M.Stat} \rightarrow \text{Inc}}$} & \footnotesize{$0.361  (0.004)$}  \\ \hline
\end{tabular}
\vspace*{-0.2cm}
\end{table}
Our aim is to check how (i.e\ through which paths) being white has influenced our model's prediction of a 0.53 probability of high income for this individual, which is approximately 30 points higher than the average probability in the cohort. Results are shown in Table \ref{tab:results_uci}, with mean and standard deviation computed by subsampling. PWSHAP shows that the local mediating effect of race through marital status is predominant. Specifying marital status increases the \emph{effect} of race by 0.361, as modeled by our black-box. Causal Shapley results in negligible direct and indirect effects, however we can't readily interpret these quantities obtained from averaging over many coalitions. This example illustrates three key attributes of PWSHAP compared with Causal Shapley, but also traditional Shapley methods overall: higher resolution, better interpretability of the resulting attributions, true locality of our explanations. The latter characteristic may explain why the effect of moderation by marital status is not captured by Causal Shapley. Here, our individual is an outlier as a large portion of white divorced individuals in the cohort have non-zero capital gain. Both the indirect and direct parts of Causal Shapley are a sum over all coalitions. This implies that for a majority of terms in Causal Shapley race and/or marital status are marginalised over, instead of being set to their feature values. The local effect of race via marital status is thus most likely ``blurred'' amongst all the coalitions that are ``causally redundant'' w.r.t. race i.e.\ where we add/drop features that are independent of race. By contrast, PWSHAP conditions on features that are independent of race (e.g. occupation), which ultimately increases the locality of our method. 

\vspace*{-0.2cm}
\section{Discussion}

PWSHAP shows how Shapley values can generate granular causal explanations for local treatment effects under a posited causal graph. Our results on both simulated and real data show the applicability and locality of our method. The interpretability and safety of PWSHAP make it a strong candidate method for evaluating algorithms in sensitive environments, assuming treatment is binary and a known cause of the outcome. Practitioners could use PWSHAP explanations to identify inequity or unfairness, prior to deployment. However, our method heavily relies on the cumbersome estimation of the conditional distributions and on the assumed statistical dependencies. Further, dividing by propensity weights can bias the results if the weights are close to zero, and our approach is limited to binary treatments. Following \citeMain{chen2018shapley}, a potential extension of our work may acknowledge the proximity of features in the DAG, instead of only considering features with a direct edge to the target variable. Deriving weights to recover the efficiency property of Shapley is another perspective for future work. PWSHAP could also be used to help guide causal discovery as it can reveal possible conditional independence relationships between variables. 
Finally, although PWSHAP is a promising alternative to address rising ethical concerns in AI, note that fairness studies rely on the availability of sensitive data, which can be challenging for practical, ethical or legal reasons \citeMain{custers2010data}. We further caution against relying exclusively on XAI when causal knowledge is insufficient or for black-box without high-accuracy, as misleading interpretations may have negative social impact. 

\section*{Acknowledgements}
We thank Shahine Bouabid, Jake Fawkes, Siu Lun Chau, Robert Hu and our anonymous reviewers for their helpful feedback. LTM and OC are supported by the EPSRC Centre for Doctoral
Training in Modern Statistics and Statistical Machine Learning (EP/S023151/1). LTM receives funding from EPSRC, OC from Novo Nordisk. KDO is funded by a Royal Society-Wellcome Trust Sir Henry Dale fellowship,
grant number 218554/Z/19/Z. CH was supported by the EPSRC Bayes4Health programme grant and The Alan Turing Institute, UK.


\newpage

\bibliographyMain{bibliography}
\bibliographystyleMain{icml2023}

\newpage
\appendix
\onecolumn


%
%




\section{Shapley Values: Definitions}
\label{sec:shapley_def}


\paragraph{Off-manifold Shapley values}
\begin{align*}
    \phi_j^\blackbox(x) 
    &=   \sum_{i=0}^{\numfeats-1}   \frac{1}{\numfeats\binom{\numfeats-1}{i}} \sum_{\substack{\includedfeats \not\owns j \\|S|=i}} [\expect[\blackbox({\localobs_{S \cup \{j\}}, \varimputed_{\overline{S \cup \{j\}}}})] - \expect[\blackbox({\localobs_{\includedfeats}, \varimputed_{\droppedfeats}})]]
\end{align*}

\paragraph{On-manifold Shapley values}
\begin{align*}
    \phi_j^\blackbox(x) 
    &=   \sum_{i=0}^{\numfeats-1}   \frac{1}{\numfeats\binom{\numfeats-1}{i}} \sum_{\substack{\includedfeats \not\owns j \\|S|=i}} [\expect[\blackbox({\localobs_{S \cup \{j\}}, \varimputed_{\overline{S \cup \{j\}}}}) \mid X_{S \cup \{j\}}=x_{S \cup \{j\}}] - \expect[\blackbox({\localobs_{\includedfeats}, \varimputed_{\droppedfeats}})]\mid X_{S}=x_{S}]
\end{align*}

\paragraph{Causal Shapley values}

\begin{align*}
    \phi_j^\blackbox(x) 
    &=   \sum_{i=0}^{\numfeats-1}   \frac{1}{\numfeats\binom{\numfeats-1}{i}} \sum_{\substack{\includedfeats \not\owns j \\|S|=i}} [\expect[\blackbox({\localobs_{S \cup \{j\}}, \varimputed_{\overline{S \cup \{j\}}}}) \mid \text{do}(X_{S \cup \{j\}}=x_{S \cup \{j\}})] - \expect[\blackbox({\localobs_{\includedfeats}, \varimputed_{\droppedfeats}})]\mid \text{do}(X_{S}=x_{S})]
\end{align*}

where the contribution of feature $j$, $\coalitionshapley{f}{S}{j}{x}$,  is decomposed into a direct and an indirect effect as follows:

\begin{align*}
\coalitionshapley{f}{S}{j}{x}
    &=  \expect[\blackbox({\localobs_{S \cup \{j\}}, \varimputed_{\overline{S \cup \{j\}}}}) \mid \text{do}(X_{S \cup \{j\}}=x_{S \cup \{j\}}) - \expect[\blackbox({\localobs_{\includedfeats}, \varimputed_{\droppedfeats}})]\mid \text{do}(X_{S}=x_{S})] \\
    &=  \expect[\blackbox({\localobs_{S \cup \{j\}}, \varimputed_{\overline{S \cup \{j\}}}}) \mid \text{do}(X_{S}=x_{S})] - \expect[\blackbox({\localobs_{\includedfeats}, \varimputed_{\droppedfeats}})]\mid \text{do}(X_{S}=x_{S})] \\
&+   \expect[\blackbox({\localobs_{S \cup \{j\}}, \varimputed_{\overline{S \cup \{j\}}}}) \mid \text{do}(X_{S \cup \{j\}}=x_{S \cup \{j\}})] - \expect[\blackbox({\localobs_{S \cup \{j\}}, \varimputed_{\overline{S \cup \{j\}}}}) \mid \text{do}(X_{S}=x_{S})]  
\end{align*}
where, in the last equality, the first line is the direct effect and the second line the indirect effect. See Section \ref{subsec:causal_shap} for details. Note that ``interventional'' Shapley values in Equation 3 of \citeMain{chen2020true} are not causal Shapley values but actually off-manifold Shapley values : the do-operator is written in Equation 3, but it is misleading as the do-operator is said to intervene ``by breaking the dependence between features in [the coalition] and the remaining features'', i.e. making the expectation marginal as in off-manifold Shapley values.

\section{Causal Inference Background}
\label{sec:ci}

\paragraph{Confounder}{A confounder is a variable that is associated with both the exposure and the outcome, causing a spurious correlation. For instance, summer is associated with eating ice cream and getting sunburns, but there is no causal relationship between the two.}
\paragraph{Mediator}{A mediator is a variable that is both an effect of the exposure and a cause of the outcome. In presence of a mediator, the total effect can be broken into two parts: the direct and indirect effect.}
\paragraph{Moderator}{A mediator is a pre-exposure variable for which the causal effect is heterogeneous in subgroups.}
\paragraph{Propensity score model}{A propensity score model is a function that predicts exposure from the observed covariates. We note it $\pi^*(c) = P(T=1|C=c)$ and note $\pi$ an estimate of $\pi^*$.}
\paragraph{Potential outcome}{As defined by the Rubin causal model \citeAppendix{rubin2005causal}, a potential outcome $Y(t)$ is the value that $Y$ would take if $T$ were set by (hypothetical) intervention to the value $t$.}
\paragraph{Identification assumptions}
\begin{itemize}
    \item \textbf{No interference} For a given individual $i$, this assumption implies that $Y_i(t)$ represents the value that $Y$ would have taken for individual i if $T$ had been set to $t$ for individual $i$, i.e\ the potential value of $Y_i$ if $T_i$ had been set to $t$.
    \item \textbf{Consistency} For a given individual $i$, $T_{i}=t \Rightarrow Y_{i}=Y_{i}(t)$. This means that for individuals who actually received exposure level $t$, their observed outcome is the same as what it would have been had they received exposure level $t$ via an hypothetical intervention. 
    \item \textbf{Conditional exchangeability} For a given individual $i$, we assume that conditional on $C$, the actual exposure level $T$ is independent of each of the potential outcomes: \\
    $Y(t) \perp T \mid \mathbf{C}, \forall t$
\end{itemize}
\textbf{Average Treatment Effect (ATE)}{The Average Treatment Effect for a binary treatment is the average difference in potential outcomes: $\mathbb{E}[Y(1)-Y(0)]$}.

\textbf{Conditional Average Treatment Effect (CATE)}{The Conditional Average Treatment Effect for a binary treatment, conditioned on $C$ is the average difference in potential outcomes: $\mathbb{E}[Y(1)-Y(0)|C=c]$. If $C$ is a sufficient adjustment set, i.e. conditional exchangeability w.r.t.~$C$ holds then the CATE can be identified as $\mathbb{E}[Y|T=1,C=c]-\mathbb{E}[Y|T=0,C=c]$.}
\paragraph{Controlled Direct Effect}{Let $Y(t,m)$ be the potential outcome under exposure level $T = t$ and mediator level $M = m$. The controlled direct effect of $T$ on outcome $Y$ comparing $T=t$ with $T=t^{*}$ and setting $M$ to $m$ measures the effect of $T$ on $Y$ not mediated through $M$ i.e.\ the effect of $T$ on $Y$ after intervening to fix the mediator to some value $m$. The controlled direct effect for individual $i$ is then $\operatorname{CDE}_i(t, t^*, m)=Y_i(t,m)-Y_i(t^{*},m)$ \citeAppendix{VanderWeele2009ConceptualIC}}
\paragraph{Natural Direct Effect}{The natural direct effect is defined as the difference between the value of the counterfactual outcome if the individual were exposed to $T = t$ and the value of the counterfactual outcome if the same individual were instead exposed to $T = t^*$, with the mediator $M$ taking whatever value it would have taken at the reference value of the exposure $T = t^*$: $Y(t,M(t^*)) – Y(t^*,M(t^*))$ \citeAppendix{VanderWeele2009ConceptualIC}}
\paragraph{Natural Indirect Effect}{The natural indirect effect is the difference, having set the exposure to a fixed level  $T = t$, between the value of the counterfactual outcome if the mediator $M$ took whatever value it would have taken at a level of the exposure $T = t$ and the value of the counterfactual outcome if the mediator assumed whatever value it would have taken at a reference level of the exposure $T = t^*$: $Y(t, M(t)) – Y(t, M(t^*))$ \citeAppendix{VanderWeele2009ConceptualIC}}

\section{Notations} \label{sec:notations}

\begin{itemize}
    \item Coalition-wise Shapley *values*  $\phi^{f}_{S,T}$ are the individual terms for a coalition in the weighted sum of the original definition of Shapley value, hence the term, *value*. It is common to use $\phi$ -even if specific to a coalition S- in reference to Shapley values. 
    \item Coalition-wise Shapley *effects* $\effect_{T \rightarrow Y | C_{S}}$ are the causal *effects* identified in  coalition-wise Shapley *values* after dividing by the propensity weights. We use $\effect$ for effects, and describe the effect of $T$ on $Y$ along the multiple paths through the covariates $C_S$. We symbolise this by using the subscript $T \rightarrow Y | C_{S}$.
    \item Path-wise Shapley *effects* $\effect_{C_i}$ are similar to coalition-wise *effects* since they are also obtained after dividing by the weights. However, the conditioning is only on the features on a single causal pathway. We thus still use $\effect$ as it is an effect, but show that the conditioning only bears upon a path using the subscript $C_i$.
\end{itemize}

\section{Further Related Work} \label{sec:further}

\subsection{Shapley Values: Axioms} \label{subsec:shap_axioms}

In this section, in line with Section \ref{sec:shapley_intro}, the Shapley value $\phi^f_{j}(x)$ is always taken with respect to value function $v$, unless specified otherwise. In the latter case, if the value function is $v'$, then the Shapley value is noted $\phi^f_{j}(x; v')$ instead. Shapley values have been shown to satisfy the four following axioms. 

\paragraph{Dummy:}{A feature $j$ receives a zero attribution if it has no possible contribution, i.e. $v_f(S \cup \{j\},x)=v_f(S,x)$ for all $S \subseteq\{1, \ldots, m\}$.}
\paragraph{Symmetry:}{Two features that always have the same contribution receive equal attribution, i.e. $v_f(S \cup \{i\},x)=v(S \cup \{j\},x)$ for all $S$ not containing $i$ or $j$ then $\phi^f_{i}(x)=\phi^f_{j}(x)$.}
\paragraph{Efficiency:} {The attributions of all features sum to the total value of all features. Formally, $\sum_{j} \phi^f_{j}(x)=v_f(\{1, . ., m\},x)$.}
\paragraph{Linearity:} {For any value function $v$ that is a linear combination of two other value functions $u$ and $w$ (i.e. $v(S)=\alpha u(S)+\beta w(S)$ ), the Shapley values of $v$ are equal to the corresponding linear combination of the Shapley values of $u$ and $w$ (i.e. $\left.\phi_{i}^f(x;v)=\alpha \phi_{i}^f(x;u)+\beta \phi_{i}^f(x;w)\right)$.}

\subsection{Causal Shapley Values} \label{subsec:causal_shap}

Heskes et. al introduced the Causal Shapley values in 2020 \citeAppendix{heskes2020causal}. For a coalition $S$, the contribution of feature $j$ $\coalitionshapley{f}{S}{j}{x}$  is decomposed into a direct and an indirect effect:
\begin{align*}
\coalitionshapley{f}{S}{j}{x}
    &=  \expect[\blackbox({\localobs_{S \cup \{j\}}, \varimputed_{\overline{S \cup \{j\}}}}) \mid \text{do}(X_{S \cup \{j\}}=x_{S \cup \{j\}}) - \expect[\blackbox({\localobs_{\includedfeats}, \varimputed_{\droppedfeats}})]\mid \text{do}(X_{S}=x_{S})] \\
    &=  \expect[\blackbox({\localobs_{S \cup \{j\}}, \varimputed_{\overline{S \cup \{j\}}}}) \mid \text{do}(X_{S}=x_{S})] - \expect[\blackbox({\localobs_{\includedfeats}, \varimputed_{\droppedfeats}})]\mid \text{do}(X_{S}=x_{S})] \\
&+   \expect[\blackbox({\localobs_{S \cup \{j\}}, \varimputed_{\overline{S \cup \{j\}}}}) \mid \text{do}(X_{S \cup \{j\}}=x_{S \cup \{j\}})] - \expect[\blackbox({\localobs_{S \cup \{j\}}, \varimputed_{\overline{S \cup \{j\}}}}) \mid \text{do}(X_{S}=x_{S})]  
\end{align*}
where, in the last equality, the first line is the direct effect and the second line the indirect effect. The direct effect measures the expected change in prediction when the stochastic feature $X_{j}$ is replaced by its feature value $x_{j}$, without changing the distribution of the other 'out-of-coalition' features. The indirect effect measures the difference in expectation when the distribution of the other 'out-of-coalition' features changes due to the additional intervention $d o\left(X_{j}=x_{j}\right)$. The direct and indirect parts of Shapley values are then be computed by taking a, possibly weighted, average over all coalitions. 

We note that in the problem setup of Section \ref{subsec:problem_Setup}
, if all covariates are pre-treatment then under mild assumptions the indirect effect of the treatment will be zero, as outlined in the following Proposition.

\begin{property}[Indirect part of Causal Shapley] \label{prop:indirect_part} Let $S$ be a coalition containing pre-treatment covariates only. We assume that an unobserved (latent) variable generates all pre-treatment covariates. Then the indirect part of the Causal Shapley values of an exposure is null, i.e.~we have
\begin{align}
    \mathbb{E}\left[f\left({C}_{\bar{S}}, c_{S}, t\right) \mid \Do({C}_{S}=c_S, T=t)\right]-\mathbb{E}\left[f\left(C_{\bar{S}}, c_{S}, t\right) \mid \Do({C}_{S}=c_{S})\right]  = 0.
\end{align}
\end{property}

The proof can be found in Supplement  \ref{proof:indirect_part}.
\subsection{Edge-Based/Flow-Based Approaches to Shapley Values} 
\label{subsec:further_graph_based}
Given that the proposed approach is model-agnostic, in this paragraph we will not review model-specific approaches that are considered to be "bespoke in nature and do not solve the problem of explainability in general" \citeAppendix{frye2019asymmetric}. Similarly, we do not review methods that violate \textit{implementation invariance}\footnote{Implementation invariance imposes that two black-box models that compute the same mathematical function have identical attributions for all features, regardless of how being implemented differently.}. Pan et. al \citeAppendix{pan2021explaining} leverage Shapley values computation to define a new quantity that distributes credit for model disparity amongst the paths in a causal graph. However, the resulting quantity isn't a Shapley value itself. Shapley Flow (SF) assigns credits to "sink-to-node" paths. To do so, SF only considers orderings that are consistent with a depth first search. Furthermore, SF modifies the original definition of Shapley values by only explaining within successive cuts of the graphs or "explanation boundaries". Such cuts are considered as alternative models to be explained. Given this modification, it is unclear what connection the explanations generated by SF exhibit with the overall model. Recursive Shapley \citeAppendix{singal2021flow} is an edge-based approach which only considers active edges. Although it provides useful insights for mediation analysis, this method overlooks the impact of confounders. Ultimately, unlike our approach both Shapley Flow and Recursive Shapley aren't additive methods, but instead hold the property of "flow conservation" which allows a parent node to split its credit amongst its children. In contrast, the \textit{efficiency} axiom of Shapley values ensure that the attributions of all features sum up to the model outcome $\blackbox(x)$. We argue that Shapley efficiency is more relevant in a regression setting, whereas flow conservation should be used for analysis of data with intrinsic ordering.

\subsection{Other Causal Approaches to Interpreting Black-Box Models} 
\label{subsec:further_causal}
Explaining black-box models in a causal manner remains challenging to this day. Zhao et. al \citeAppendix{zhao2021causal} expand on the use of partial dependence plot (PDP), where the dependence on a set of covariates is computed by taking the expectation of the model over the marginal distribution of all other covariates. They note that the PDP formula is similar to Pearl’s back-door adjustment. More specifically, marrying Shapley values with causal reasoning has been an active research question. Janzing et. al \citeAppendix{janzing2020feature} considers the model’s prediction process itself to be a causal process: from features to model inputs and ultimately model output. The authors claim that marginal Shapley values can be apprehended in terms of \textit{do}-calculus, if we consider that setting a feature to a given value is equivalent to intervening on it. As such, marginal or so-called off-manifold Shapley values may be sufficient to explain that specific causal process but this approach is contrived as it does not consider the real-world causal relationships between features. This approach however does not acknowledge any underlying causal structure from the real world. New causality based formulations of Shapley values have been proposed to compute feature attributions from a hypothesised causal structure of the data. Asymmetric Shapley values \citeAppendix{frye2019asymmetric} use conditioning by observation but only consider causally-consistent coalitions i.e coalitions such that known causal ancestors precede their descendants. The resulting explanations quantify the impact a given feature has on model prediction while its descendants remain unspecified. As a result, they ignore downstream effects in favour of root causes \citeAppendix{wang2021shapley}. 

Below is a table showing a summary comparison of existing causality-based or graph-based Shapley approaches with PWSHAP.
Node efficiency refers to the original efficiency property of Shapley values: $\blackbox(x)=\phi^{\blackbox}_{0}(x)+\sum_{i=1}^{M} \phi^{\blackbox}_{i}(x)$, where $\phi^{\blackbox}_{j}(x)$ is the contribution of feature $j$ to $\blackbox(x)$ and $\phi^{\blackbox}_{0}(x)=\mathbb{E} [f(X))]$ is the averaged prediction with the expectation over the observed data distribution. By game at each node/within each boundary we describe the fact that Shapley Flow and Recursive Shapley consider successive cuts from the graphs. Flow conservation or cut efficiency is the equivalent efficiency property, within such a cut. We refer the reader to the corresponding papers for further details.

\begin{table}[H]
\centering
\caption{Comparison between PWSAP and existing causality-based or graph-based Shapley approaches}
\label{tab:compar}
\resizebox{\textwidth}{!}{%
\begin{tabular}{lccccc}
\cline{2-6}
                                                                               & Shapley Flow & Recursive Shapley & Asymmetric Shapley        & Causal Shapley & PWSHAP \\ \toprule
\begin{tabular}[c]{@{}l@{}}Flow conservation or \\ cut efficiency\end{tabular} & X            & X                 &                           &    &        \\
\midrule
Node efficiency                                                                &              &                   & X                         & X  & X      \\
\midrule
Node-based                                                                     &              &                   & X                         & X  & X      \\

\midrule
Edge-based                                                                     &              & X                 &                           &    &        \\
\midrule
\begin{tabular}[c]{@{}l@{}}Source-to-sink \\ path-based\end{tabular}           & X            &                   &                           &    &        \\
\midrule
Path-based                                                                     &              &                   &                           &    & X      \\
\midrule
\begin{tabular}[c]{@{}l@{}}Game at each node or \\ within each boundary \\ of explanation\end{tabular} & X & X &  &  &  \\
\midrule
Ignores direct effects                                                         &              &                   & X                         &    &        \\
\midrule
Fidelity to original Shapley                                                   &              &                   & \begin{tabular}[c]{@{}c@{}}X \\ (but violates symmetry axiom)\end{tabular} & X  & X      \\ \bottomrule
\end{tabular}%
}
\end{table}

Ultimately, in a previous work \citeAppendix{sani2020explaining}, Sani et. al  (2020) introduce an XAI method which uses auxiliary interpretable labels that are assumed to be readily available. Although their method has shown to perform well, it relies on the latter strong assumption which limits its applicability. Note that, like in PWSHAP, this approach further requires external information (besides the model and input/output data). We believe Sani et. al's method can be complementary to ours. For instance, in settings where the causal quantity of interest is identified from the obtained Partial Ancestral Graph, one may train a model regressing predictions on the interpretable labels and apply PWSHAP to the resulting model. We thank our anonymous reviewers for this remark.

\subsection{Further discussion on an XAI model's dependency to the DAG} 
\label{subsec:further_dag}

We view the fact that our approach is agnostic to the choice of (compatible) DAG as a strength, as it allows different experts to explain the black-box model output according to their own causal beliefs about the data or phenomenon being studied.
For example, take the classical DAG with confounders, a treatment and an outcome (in Figure \ref{fig:dag_blocks_infography}, graph 2). The black-box model takes all confounders and the treatment as inputs. However it does not ``know'' whether the covariates are actually confounders or mediators, and generates its output regardless. In PWSHAP, given that we have the additional (external) knowledge of the DAG representing our beliefs, we would interpret the output as a conditional average treatment effect. More generally, a model only gives indications about causal quantities that are already identified with statistical quantities according to prior causal assumptions, as in the causal roadmap by Petersen and van der Laan (2014) \citeAppendix{petersen2014causal}). Although causal discovery using both the model outputs \textit{and the data} can yield a set of candidate DAGs that share the same edges, in the form of a Partial Ancestral Graph (PAG), some directions may be missing and one may still need to choose a DAG amongst multiple alternatives.

\subsection{Bias, Mediation and Moderation Analysis}
\label{sec:other_analyses}

Our approach has added value compared to existing methods for sensitivity, mediation and moderation analysis. In the following paragraph, we review the state of the art with regards to each of these objectives. \\

\paragraph{Moderation analysis} A common approach to assess moderation is to (i) fit an Heterogeneous Treatment Effect (HTE) model that predicts an individual's treatment effect from a set of covariates (ii) find subgroups with similar treatment effects. Subgroups can be infered directly from the data \citeAppendix{imai2013estimating, wang2017causal}, from the individual predicted treatment effects \citeAppendix{foster2011subgroup} or using statistical hypothesis tests \citeAppendix{athey2016recursive, song2007method, holmes2018machine}. The main drawback of subgroup findings methods is that they are prone under-powered and time-consuming. Holmes et. al \citeAppendix{holmes2018machine} introduce a partitioning method which controls the type I error, however it is still limited to comparing subgroups two by two. Another approach to moderation analysis is to use interpretable models to predict HTEs. Nilsson et. al \citeAppendix{nilsson2019assessing} build two potential outcomes models (treated/untreated) and fit a regression model to predict their difference from the covariates of interest. Regression coefficients are ultimately used as a measure of moderation, but this solution is prone to model misspecification. Explanation methods, and in particular feature attribution models allow for a finer-grained understanding of the sources of heterogeneity. More recently, Wu et. al suggested to use Distillation to generate explanations of HTE models and assess moderation induced by each covariate \citeAppendix{wu2021distilling}. However, explanation models such as Distillation that involve building a simpler surrogate model have received criticism for \textit{approximating} the target black-box function instead of explaining it \citeAppendix{rudin2019stop}. Ultimately, HTE models are built without taking the causal structure into consideration the causal structure e.g. not conditioning on post-treatment features. To the best of our knowledge, our approach is the first method that can assign attribution to moderators directly from the outcome regression model whilst acknowledging the posited causal structure and the rules of \textit{do}-calculus by Pearl.

\paragraph{Sensitivity analysis}

In most treatment effect estimation studies, it is assumed that all confounders of treatment and outcome are observed. This is a strong assumption and one might wonder whether results will be greatly perturbed or not by the presence of an unobserved confounder. Sensitivity analysis generally aims at determining what the impact of a given amount of unobserved confounding would be on causal conclusions of the study. In particular, as we have no access to the unobserved confounder, we make assumptions about its relationship with treatment and outcome. One line of work is to assume parameters for this relationship and infer the rest of the model when values of these parameters are fixed, e.g.~via maximum likelihood \citeAppendix{veitch2020sense, rosenbaum1983assessing, imbens2003sensitivity}. As a result, one can check the change in treatment effects with fixed values of the unobserved confounder \citeAppendix{rosenbaum1983assessing} or draw contour plots showing the bias depending on the parameters \citeAppendix{veitch2020sense, imbens2003sensitivity}. Another line of work assumes a fixed ratio between the propensity score with only observed covariates and a variation of the propensity score that also includes unobserved confounders \citeAppendix{rosenbaum2005sensitivity, tan2006distributional, jesson2022scalable}. This ratio quantifies how much hidden confounding is present (with a ratio of 1 being no hidden confounding) and is set by the user. As a result, one can deduce intervals for inference quantities like p-values or treatment effects from a given ratio. This can be leveraged to find the lowest ratio that makes the interval reach thresholds invalidating causal conclusions, e.g.~0.05 for a p-value or zero for the treatment effects. The higher the ratio has to be, the more robust to unobserved confounding the study. This idea is close to the E-value, a scalar metric representing the minimal amount of unobserved confounding needed to fully explain away the treatment-outcone relationship \citeAppendix{vanderweele2017sensitivity}. Although sensitivity analysis can also be applied to assess the role of a given observed confounder \citeAppendix{veitch2020sense, imbens2003sensitivity}, it remains different from our local bias analysis approach that only applies to observed confounders. However, unlike most sensitivity analysis methods, our approach does not rely on parameters or assumptions other than the joint distribution of the data, and it summarises the confounding of a given pre-treatment covariate in a single bias scalar.

\paragraph{Mediation analysis}\label{sec:med_analysis}

The state-of-the-art approach to mediation analysis is based on Natural Direct and Indirect Effects \citeAppendix{VanderWeele2009ConceptualIC}. Computation of Natural Direct/Indirect Effects requires four assumptions \citeAppendix{VanderWeele2009ConceptualIC} : 1) no unmeasured confounding for the exposure-outcome relationship, 2) no unmeasured confounding for the mediator-outcome relationship, 3) no unmeasured confounding for the exposure-mediator relationship, 4) no mediator-outcome confounding that is itself affected by the exposure. These are strong hypotheses, with the latter typically being considered to be unrealistic.
PWSHAP also implicitly relies on these assumptions.
However, it is common to assume unconfoundedness of the exposure-outcome relationship. Indeed, the exposure is naturally or experimentally randomised in many mediation analysis settings, such as fairness studies (e.g. when sex or race are the exposure). Thus, without confounders, PWSHAP effects are causal in these settings. 
Further, our approach to mediation analysis is more faithful to causal inference than Causal Shapley. By comparing CDEs, we assess the effect of setting the given mediator to its value. By contrast, Causal Shapley breaks the relationship between treatment and mediator when intervening on the mediator in the indirect effect. A local mediation analysis example is given in Supplement \ref{subsec:local_mediation}. Supplement \ref{sec:further_blocks} details the application of PWSHAP to dependent mediators or in presence of both confounders and mediators.

\section{Detailed Results for the ``Building Blocks'' Examples} \label{sec:detailed_building_blocks}

\subsection{Local Moderation Analysis} \label{subsec:local_moderation}

\textbf{In the following, we prove the result of the first example where treatment is randomised} \\
We assume the following model
\begin{align*}
    C_1, C_2 &\sim \text{Uniform}(0,1), \ \ \ C_1 \indep C_2 \\
    Y &= \beta T + \gamma_1C_1 + \gamma_2C_2 + \alpha_1TC_1 + \alpha_2TC_2 + \epsilon
\end{align*}
with $\mathbb{E}[\epsilon|T,C_1,C_2] = 0$. Assuming the true outcome model is known, our aim is to explain the following black-box: $f^*(c_1, c_2, t) = \beta t + \gamma_1c_1 + \gamma_2c_2 + \alpha_1tc_1 + \alpha_2tc_2$. We further assume that treatment is randomised by taking $T \sim \text{Bernoulli}(p)$.

We note that, from Property \ref{prop:shapley_decomposition},
\begin{align*}
\coalitionshapley{f^*,\text{obs}}{\{C_1,C_2\}}{T}{c_1,c_2,t} = \coalitionshapley{f^*,\text{causal}}{\{C_1,C_2\}}{T}{c_1,c_2,t}  &= (t - p)(\beta + \alpha_1c_1 + \alpha_2c_2) \\
    \coalitionshapley{f^*,\text{obs}}{\{C_1\}}{T}{c_1,t} = \coalitionshapley{f^*,\text{causal}}{\{C_1\}}{T}{c_1,t} &= (t - p)(\beta + \alpha_1c_1 + \alpha_2/2) \\
    \coalitionshapley{f^*,\text{obs}}{\{C_2\}}{T}{c_2,t} = \coalitionshapley{f^*,\text{causal}}{\{C_2\}}{T}{c_2,t} &= (t - p)(\beta + \alpha_1/2 + \alpha_2c_2) \\
    \coalitionshapley{f^*,\text{obs}}{\emptyset}{T}{t} = \coalitionshapley{f^*,\text{causal}}{\emptyset}{T}{t} &= (t - p)(\beta + \alpha_1/2 + \alpha_2/2) \\ \\
\end{align*}

All of these Causal Shapley values only correspond to direct effects, as the indirect effect is zero from Property \ref{prop:indirect_part}.

\subsection{Local Bias Analysis}
\label{subsec:local_bias}

We compare PWSHAP with Causal Shapley on a bias analysis example where $C_1$ and $C_2$ are distributed as before, but we assume instead that treatment allocation depends only on one covariate $C_1$: $\mathbb{E}[T|C_1,C_2] = C_1^\alpha$ which implies that $C_1$ is both a confounder and a moderator whereas $C_2$ only acts as a moderator. PWSHAP effects are then given as:
\begin{align*}
\effect^{f^*}_{T \leftarrow C_1 \rightarrow Y, \atop C_1: T \rightarrow Y} = \alpha_1(c_1 - \frac{\alpha+1}{\alpha+2}) - \gamma_1\frac{\alpha+1}{2(\alpha+2)}             &  &\effect^{f^*}_{C_2: T \rightarrow Y} = \alpha_2(c_2 - \frac{1}{2})              \\
\effect^{f^*}_{T \rightarrow Y | \emptyset} = \beta + \gamma_1\frac{\alpha+1}{2(\alpha + 2)} + \alpha_1\frac{\alpha+1}{\alpha+2} + \frac{\alpha_2}{2}.          &  & 
\end{align*}

We note that $\mathbb{E}[\effect^{f^*}_{C_2: T \rightarrow Y}(C_1,C_2)] = 0$ but $\mathbb{E}[\effect^{f^*}_{T \leftarrow C_1 \rightarrow Y, \atop C_1: T \rightarrow Y}(C_1,C_2)] \neq 0$. This illustrates not only Lemma \ref{lemma:local_confounding_effect}, but also to the following corollary where we do not assume the model is true.

\begin{corollary}[Integration of the local confounding effect, black-box model] \label{cor:local_confounding_effect_blackbox}
 Let $C_1, C_2$ be two pre-treatment covariates such that $C_2 \indep T | C_1$. Then the integral of the local confounding effect w.r.t. $C_2$ on the joint distribution of covariates is null 
\centerline{ $ \mathbb{E}[\effect^f_{T \leftarrow C_2 \rightarrow Y}(C_1,C_2)] = 0.$ }
\end{corollary}
The proof can be found in Supplement \ref{proof:local_confounding_effect_blackbox}.
By comparison, in Causal Shapley values, only the direct part is non null: 
\begin{align*}
    &\phi^{f^*,\text{causal}}_{T,\text{direct}} = \beta \big(t - \frac{c_1^\alpha}{2} - \frac{1}{2(\alpha+1)} \big)
     + \alpha_1 \big(\frac{c_1}{2}(t - c_1^\alpha) +\frac{1}{2}( \frac{t}{2} - \frac{1}{\alpha+2}) \big) \\
     & \hspace{2cm} + \alpha_2 \big((\frac{c_2}{3} + \frac{1}{12})(t - c_1^\alpha) + (\frac{c_2}{6} + \frac{1}{6}(t - \frac{1}{\alpha+1}) \big). \\
\end{align*}

\textbf{Proof :} First let's note that

\begin{align*}
    \mathbb{E}[Y|T=1,C_1=c_1,C_2=c_2] - \mathbb{E}[Y|T=0,C_1=c_1,C_2=c_2] = \beta + \alpha_1c_1 + \alpha_2a_2
\end{align*}

Then we show that :

\begin{align*}
    &\mathbb{E}[Y|T=1,C_1=c_1] - \mathbb{E}[Y|T=0,C_1=c_1] = \beta + \alpha_1c_1 + \frac{a_2}{2} \\
    &\mathbb{E}[Y|T=1,C_2=c_2] - \mathbb{E}[Y|T=0,C_2=c_2] = \beta + \gamma_1\frac{\alpha+1}{2(\alpha+2)} + \alpha_1\frac{\alpha+1}{\alpha+2} + \alpha_2c_2 \\
    &\mathbb{E}[Y|T=1] - \mathbb{E}[Y|T=0] = \beta + \gamma_1\frac{\alpha + 1}{2(\alpha + 2)} + \alpha_1\frac{\alpha+1}{\alpha+2} + \frac{\alpha_2}{2}   \\
\end{align*}

First, let us note that, using independence of $C_1$ and $C_2$,
\begin{align*}
    &\mathbb{E}[T|C_1=c_1] =  \mathbb{E}[\mathbb{E}[T|C_1=c_1,C_2]|C_1=c_1] = \mathbb{E}[ c_1^\alpha |C_1=c_1] = c_1^\alpha \\
    &\mathbb{E}[T|C_2=c_2] =  \mathbb{E}[\mathbb{E}[T|C_2=c_2,C_1]|C_2=c_2] = \mathbb{E}[ C_1^\alpha |C_2=c_2] = \frac{1}{\alpha+1} \\
    &\mathbb{E}[T] = \frac{1}{\alpha+1}
\end{align*}

By Bayes's rule and independence of $C_1$ and $C_2$,
\begin{align*}
    p(c_2|c_1,t=1) = \frac{p(t=1|c_1,c_2)p(c_1)p(c_2)}{p(t=1|c_1)p(c_1)} = \frac{p(t=1|c_1,c_2)}{p(t=1|c_1)} = \frac{c_1^\alpha }{c_1^\alpha} = 1
\end{align*}
and, similarly,
\begin{align*}
    p(c_2|c_1,t=0) = 1
\end{align*}
As a result,
\begin{align*}
&\mathbb{E}[C_2|c_1,t=1] = \mathbb{E}[C_2|c_1,t=0] = \frac{1}{2} \\
&\mathbb{E}[C_2|c_1,t=1] - \mathbb{E}[C_2|c_1,t=0] = 0
\end{align*}
and
\begin{align*}
    \mathbb{E}[Y|T=1,C_1=c_1] - \mathbb{E}[Y|T=0,C_1=c_1] &= \beta + \gamma_2(\mathbb{E}[C_2|c_1,t=1] - \mathbb{E}[C_2|c_1,t=0]) + \alpha_1c_1 + \alpha_2\mathbb{E}[C_2|c_1,t=1] \\
    &= \beta + \alpha_1c_1 + \frac{a_2}{2} \\
\end{align*}
which proves the first equality. For the second equality, we have
\begin{align*}
    p(c_1|c_2,t=1) = \frac{p(t=1|c_1,c_2)p(c_1)p(c_2)}{p(t=1|c_2)p(c_2)} = \frac{c_1^\alpha}{\frac{1}{\alpha+1}} = (\alpha+1)c_1^\alpha
\end{align*}
and, similarly,
\begin{align*}
    p(c_1|c_2,t=0) = \frac{1 - c_1^\alpha}{1 - \frac{1}{\alpha+1}}
\end{align*}
Thereby, we obtain
\begin{align*}
&\mathbb{E}[C_1|c_2,t=1] = \frac{\alpha+1}{\alpha+2} \\
&\mathbb{E}[C_1|c_2,t=0] = \frac{\alpha+1}{2(\alpha+2)} \\
&\mathbb{E}[C_1|c_2,t=1] - \mathbb{E}[C_1|c_2,t=0] = \frac{\alpha+1}{2(\alpha+2)}
\end{align*}
and
\begin{align*}
    \mathbb{E}[Y|T=1,C_2=c_2] - \mathbb{E}[Y|T=0,C_1=c_1] &= \beta + \gamma_1(\mathbb{E}[C_1|c_2,t=1] - \mathbb{E}[C_1|c_2,t=0]) \\
    & \ \ \ \ \ \ \ \ \ \ \ \ \ \ \ \ \ \ + \alpha_2c_2 + \alpha_1\mathbb{E}[C_1|c_2,t=1] \\
    &= \beta + \gamma_1\frac{\alpha+1}{2(\alpha+2)} + \alpha_2c_2 + \alpha_1\frac{\alpha+1}{\alpha+2} \\
\end{align*}

Similarly, for the third equality, we note that, as before,
\begin{align*}
    p(c_2|t=1) &= 1 \\
    p(c_2|t=0) &= 1 \\
    p(c_1|t=1) &= (\alpha+1)c_1^\alpha \\
    p(c_1|t=0) &= \frac{1 - c_1^\alpha}{1 - \frac{1}{\alpha+1}}
\end{align*}
which leads to, as before,
\begin{align*}
&\mathbb{E}[C_2|t=1] = \frac{1}{2} \\
&\mathbb{E}[C_2|t=0] = \frac{1}{2} \\
&\mathbb{E}[C_2|t=1] - \mathbb{E}[C_2|c_1,t=0] = 0 \\
&\mathbb{E}[C_1|t=1] = \frac{\alpha+1}{\alpha+2} \\
&\mathbb{E}[C_1|t=0] = \frac{\alpha+1}{2(\alpha+2)} \\
&\mathbb{E}[C_1|t=1] - \mathbb{E}[C_1|c_2,t=0] = \frac{\alpha+1}{2(\alpha+2)}
\end{align*}
thereby
\begin{align*}
    \mathbb{E}[Y|T=1] - \mathbb{E}[Y|T=0] &= \beta + \gamma_1 (\mathbb{E}[C_1|T=1] - \mathbb{E}[C_1|T=0] ) \\
    &+ \gamma_2(\mathbb{E}[C_2|T=1] - \mathbb{E}[C_2|T=0]) \\
    &+ \alpha_1\mathbb{E}[C_1|T=1] + \alpha_2\mathbb{E}[C_2|T=1]  \\
    &= \beta + \gamma_1\frac{\alpha + 1}{2(\alpha + 2)} + \alpha_1\frac{\alpha+1}{\alpha+2} + \frac{a_2}{2}
\end{align*}

Now, for Causal Shapley values, we can show that
\begin{align*}
    \mathbb{E}[Y|\Do(t,c_1,c_2)] - \mathbb{E}[Y|\Do(c_1,c_2)] &= (t - c_1^\alpha )(\beta + \alpha_1c_1 + \alpha_2c_2) \\
    \mathbb{E}[Y|\Do(t,c_1)] - \mathbb{E}[Y|\Do(c_1)] &= \beta (t - c_1^\alpha) + \alpha_1c_1(t - c_1^\alpha) + \alpha_2(\frac{t}{2} - \frac{c_1^\alpha}{2}) \\
    \mathbb{E}[Y|\Do(t,c_2)] - \mathbb{E}[Y|\Do(c_2)] &= \beta (t - \frac{1}{\alpha+1}) + \alpha_2c_2(t - \frac{1}{\alpha+1}) + \alpha_1(\frac{t}{2} - \frac{1}{\alpha+2}) \\
    \mathbb{E}[Y|\Do(t)] - \mathbb{E}[Y] &= \beta(t - \frac{1}{\alpha+1}) + \alpha_1(\frac{t}{2} - \frac{1}{\alpha+2}) + \alpha_2(\frac{t}{2} - \frac{1}{2(\alpha+1)}) \\
\end{align*}
as $\mathbb{E}[TC_1|c_1] = c_1^{\alpha+1}$, $\mathbb{E}[TC_1|c_2] = \mathbb{E}[TC_1] = \frac{1}{\alpha+2}$, $\mathbb{E}[TC_2|c_1] = \frac{c_1^\alpha}{2}$, $\mathbb{E}[TC_2|c_2] = \frac{c_2}{\alpha+1}$, $\mathbb{E}[TC_2] = \frac{1}{2(\alpha+1)}$

\subsection{Local Mediation Analysis} \label{subsec:local_mediation}

We compare PWSHAP with Causal Shapley on a mediation analysis example inspired by the Berkeley dataset \citeAppendix{bickel1975sex}. An algorithm predicts the probability of success of an applicant to a college. In this example, $X=(T,Q,D)$ where $T$ is the gender of the applicant $Q$ is an exam result and $D$ is the department. We assume $Q \sim \text{Uniform}(0,1)$, $D | T=0 \sim \text{Bernoulli}(0.8)$ and $D | T=1 \sim \text{Bernoulli}(0.2)$. $D$ is a mediator of gender however $Q$ is only an ancestor of the outcome, and not a mediator. Our black-box is the true outcome model, and $Y = \alpha_Q Q + \alpha_D D + \alpha_T T + \alpha_{DT} DT + \alpha_{QT} QT + \epsilon$, with $\mathbb{E}[\epsilon | D, Q, T] = 0$.
\begin{align*}
    \effect^{f^*}_{T \rightarrow D \rightarrow Y} = 0.6 \alpha_D + \alpha_{DT} (d - \frac{1}{5})  & & \effect^{f^*}_{T \rightarrow Q \rightarrow Y} = \alpha_{QT}(q - \frac{1}{2}) 
\end{align*}
\begin{align*}
    \effect^{f^*}_{T \rightarrow Y | \emptyset} &= \alpha_T - 0.6 \alpha_D + \frac{\alpha_{DT}}{5} + \frac{\alpha_{QT}}{2}  \\
    \phi^{f^*,\text{causal}}_{T,\text{direct}} &= \alpha_T(t - \frac{1}{2}) + \alpha_{DT}[\frac{d}{2}(t - \frac{1}{2}) + \frac{1}{2}(\frac{t}{2} - \frac{1}{10})] + \frac{\alpha_{QT}}{2}(t - \frac{1}{2})(q + \frac{1}{2}) \\
    \phi^{f^*,\text{causal}}_{T,\text{indirect}} &= \frac{\alpha_D}{2}  (\frac{3}{10} - \frac{3t}{5}) \\
\end{align*}
We note that $\int_{q} \effect^{f^*}_{T \rightarrow Q \rightarrow Y}(p,q) dp(q) = 0$ but $\int_{d} \effect^{f^*}_{T \rightarrow D \rightarrow Y}(p,q) dp(d) \neq 0$.

\textbf{Proof :} Coalition-wise Shapley effects are :
\begin{align*}
     \effect^{f^*}_{T \rightarrow Y | D, Q}(d,q)
     &= \text{CDE}(d,q) \\
     &= \mathbb{E}[Y|T=1,D=d,Q=q] - \mathbb{E}[Y|T=0,D=d,Q=q] \\ 
     &= \alpha_T + \alpha_{DT}d + \alpha_{QT}q \\
    \effect^{f^*}_{T \rightarrow Y | D}(d)
     &= \text{CDE}(d) \\
     &= \mathbb{E}[Y|T=1,D=d] - \mathbb{E}[Y|T=0,D=d] \\ 
     &= \alpha_T + \alpha_{DT}d + \alpha_{QT}\mathbb{E}[Q|T=1] \\
     &= \alpha_T + \alpha_{DT}d + \alpha_{QT}\mathbb{E}[Q] \\
     &= \alpha_T + \alpha_{DT}d + \alpha_{QT}\frac{1}{2} \\
    \effect^{f^*}_{T \rightarrow Y | Q}(q)
     &= \text{CDE}(q) \\
     &= \mathbb{E}[Y|T=1,Q=q] - \mathbb{E}[Y|T=0,Q=q] \\ 
     &= \alpha_T + \alpha_{DT}\mathbb{E}[D|T=1] + \alpha_{QT}q + \alpha_D(\mathbb{E}[D|T=1] - \mathbb{E}[D|T=0]) \\
     &= \alpha_T + \frac{\alpha_{DT}}{5} + \alpha_{QT}q - \alpha_D\frac{3}{5} \\
    \effect^{f^*}_{T \rightarrow Y | \emptyset}(q)
     &= \text{ATE} \\
     &= \mathbb{E}[Y|T=1] - \mathbb{E}[Y|T=0] \\ 
     &= \alpha_T + \alpha_{DT}\mathbb{E}[D|T=1] + \alpha_{QT}\mathbb{E}[Q|T=1] + \alpha_D(\mathbb{E}[D|T=1] - \mathbb{E}[D|T=0]) \\
     &= \alpha_T + \frac{\alpha_{DT}}{5} + \frac{\alpha_{QT}}{2} - \alpha_D\frac{3}{5} \\
\end{align*}

As a result, we deduce path-wise effects
\begin{align*}
    &\effect^{f^*}_{T \rightarrow D \rightarrow Y} = \effect^{f^*}_{T \rightarrow Y | D, Q} - \effect^{f^*}_{T \rightarrow Y | Q} = 0.6 \alpha_D + \alpha_{DT} (d - \frac{1}{5}) \\
    &\effect^{f^*}_{T \rightarrow Q \rightarrow Y} = \effect^{f^*}_{T \rightarrow Y | D, Q} - \effect^{f^*}_{T \rightarrow Y | D} = \alpha_{QT}(q - \frac{1}{2}) \\
\end{align*}

Causal Shapley values are :
\begin{align*}
    \coalitionshapley{f^*,\text{causal}}{\{D,Q\},\text{direct}}{T}{d,q,t} &= (\alpha_T + \alpha_{DT} d + \alpha_{QT} q)(t - \frac{1}{2} ) \\
    &= \mathbb{E}[f(d,q,t) | \Do(d,q)] - \mathbb{E}[f(d,q,T) | \Do(d,q)] \\
    &= \alpha_T(t - \mathbb{E}[T]) + \alpha_{DT}(dt - d\mathbb{E}[T]) + \alpha_{QT}(qt - q\mathbb{E}[T]) \\
    &= (t - \frac{1}{2})(\alpha_T + \alpha_{DT}d + \alpha_{QT}q) \\
    \coalitionshapley{f^*,\text{causal}}{\{D,Q\},\text{indirect}}{T}{d,q,t} 
    &=  \mathbb{E}[f(d,q,t) | \Do(d,q,t)] - \mathbb{E}[f(d,q,t) | \Do(d,q)] \\
    &= 0 \\
    \coalitionshapley{f^*,\text{causal}}{\{D\},\text{direct}}{T}{d,t}
    &= \mathbb{E}[f(d,Q,t) | \Do(d)] - \mathbb{E}[f(d,Q,T) | \Do(d)] \\
    &= \alpha_T(t - \mathbb{E}[T]) + \alpha_{DT}(dt - d\mathbb{E}[T]) + \alpha_{QT}(\mathbb{E}[Q]t - \mathbb{E}[QT]) \\
    &= (\alpha_T + \alpha_{DT} d + \frac{\alpha_{QT}}{2} )(t - \frac{1}{2} ) \\
    \coalitionshapley{f^*,\text{causal}}{\{D\},\text{indirect}}{T}{d,t}
    &=  \mathbb{E}[f(d,Q,t) | \Do(d,t)] - \mathbb{E}[f(d,Q,t) | \Do(d)] \\
    &= 0 \\
    \coalitionshapley{f^*,\text{causal}}{\{Q\},\text{direct}}{T}{q,t}
    &= \mathbb{E}[f(D,q,t) | \Do(q)] - \mathbb{E}[f(D,q,T) | \Do(q)] \\
    &= \alpha_T(t - \mathbb{E}[T]) + \alpha_{DT}(\mathbb{E}[D]t - \mathbb{E}[DT]) + \alpha_{QT}(qt - q\mathbb{E}[T]) \\
    &=  \alpha_T(t - \frac{1}{2}) +  \alpha_{DT}(\frac{t}{2} - \frac{1}{10}) + \alpha_{QT}q (t - \frac{1}{2}) \\
    \coalitionshapley{f^*,\text{causal}}{\{Q\},\text{indirect}}{T}{q,t}  \\
    &=  \mathbb{E}[f(D,q,t) | \Do(q,t)] - \mathbb{E}[f(D,q,t) | \Do(q)] \\
    &= \alpha_D (\mathbb{E}[D|t] - \mathbb{E}[D]) \\
    &= \alpha_D (\frac{3}{10} - \frac{3t}{5}) \\
    \coalitionshapley{f^*,\text{causal}}{\emptyset,\text{direct}}{T}{t}
    &= \mathbb{E}[f(D,Q,t)] - \mathbb{E}[f(D,Q,T)] \\
    &= \alpha_T(t - \mathbb{E}[T]) + \alpha_{DT}(\mathbb{E}[D]t - \mathbb{E}[DT]) + \alpha_{QT}(\mathbb{E}[Q]t - q\mathbb{E}[QT]) \\
    &=  \alpha_T(t - \frac{1}{2}) + \alpha_{DT}(\frac{t}{2} - \frac{1}{10}) + \frac{\alpha_{QT}}{2} (t - \frac{1}{2}) \\
    \coalitionshapley{f^*,\text{causal}}{\emptyset,\text{indirect}}{T}{t} \\
    &=  \mathbb{E}[f(D,Q,t) | \Do(t)] - \mathbb{E}[f(D,Q,t)] \\
    &= \alpha_D (\mathbb{E}[D|t] - \mathbb{E}[D]) \\
    &= \alpha_D (\frac{3}{10} - \frac{3t}{5}) \\
\end{align*}
Summing them altogether with appropriate binomial weights, we have

\begin{align*}
    \phi^{f^*,\text{causal}}_{T,\text{direct}} &= \alpha_T(t - \frac{1}{2}) + \alpha_{DT}[\frac{d}{2}(t - \frac{1}{2}) + \frac{1}{2}(\frac{t}{2} - \frac{1}{10})] + \frac{\alpha_{QT}}{2}(t - \frac{1}{2})(q + \frac{1}{2}) \\
    \phi^{f^*,\text{causal}}_{T,\text{indirect}} &= \frac{\alpha_D}{2}  (\frac{3}{10} - \frac{3t}{5}) \\
\end{align*}

\section{Further Results for Complex ``Building Blocks'' DAGS} \label{sec:further_blocks}

\subsection{A Mix of Confounders and Mediators}

We now assume the model :
\begin{align*}
    C_1 &\sim \text{Bernoulli}(\frac{1}{2}) \\
    C_2 &\sim \text{Bernoulli}(\frac{1}{2}) \\
    Q | C_1 &\sim \text{Bernoulli}(1 - C_1) \\
    T | C_1 &\sim \text{Bernoulli}(C_1) \\
    D | T,C_1 &\sim \text{Bernoulli}(\frac{4}{5} - \frac{3}{5}\frac{T+C_1}{2}) \\
    Y &= \alpha_Q Q + \alpha_D D + \alpha_T T + \alpha_1 C_1 + \alpha_2 C_2 + \alpha_{DT} DT + \alpha_{QT} QT \\
    & \ \ \ \ \ + \alpha_{1T} C_1T + \alpha_{2T} C_2T   + \epsilon, \text{ with } \mathbb{E}[\epsilon | D, Q, T] = 0.
\end{align*}
We have,
\begin{align*}
    \effect^{f^*}_{T \rightarrow Y | C_1, C_2, D, Q}(c_1,c_2,d,q) &= \alpha_T + \alpha_{DT}d + \alpha_{QT}q + \alpha_{1T}c_1 + \alpha_{2T}c_2 \\
    \effect^{f^*}_{T \rightarrow Y | C_1, C_2, Q}(c_1,c_2,q) &= \alpha_T + \alpha_{DT}\mathbb{E}[D|T=1,c_1,c_2,q] + \alpha_{QT}q + \alpha_{1T}c_1 + \alpha_{2T}c_2 \\
    & \ \ \ \ \ + \alpha_D (\mathbb{E}[D|T=1,c_1,c_2,q] - \mathbb{E}[D|T=0,c_1,c_2,q]) \\
    &= \alpha_T + \alpha_{DT}(\frac{1}{2} - \frac{3c_1}{10}) + \alpha_{QT}q + \alpha_{1T}c_1 + \alpha_{2T}c_2 - \frac{3\alpha_D}{10} \\
    \effect^{f^*}_{T \rightarrow Y | C_1, C_2, D}(c_1,c_2,d) &= \alpha_T + \alpha_{DT}d + \alpha_{QT}\mathbb{E}[D|T=1,c_1,c_2,d] + \alpha_{1T}c_1 + \alpha_{2T}c_2 \\
    & \ \ \ \ \ + \alpha_Q (\mathbb{E}[Q|T=1,c_1,c_2,d] -  \mathbb{E}[Q|T=0,c_1,c_2,d]) \\
    &= \alpha_T + \alpha_{DT}d + \alpha_{QT}(1 - c_1) + \alpha_{1T}c_1 + \alpha_{2T}c_2 \\
    \effect^{f^*}_{T \rightarrow Y | C_1, D, Q}(c_1,d,q) &= \alpha_T + \alpha_{DT}d + \alpha_{QT}q + \alpha_{1T}c_1 + \frac{\alpha_{2T}}{2} \\
    \effect^{f^*}_{T \rightarrow Y | C_2, D, Q}(c_2,d,q) &= \alpha_T + \alpha_{DT}d + \alpha_{QT}q + \alpha_{2T}c_2 + \alpha_{1T}\mathbb{E}[C_1|T=1,d,q]\\
    & \ \ \ \ \ + \alpha_1(\mathbb{E}[C_1|T=1,d,q] - \mathbb{E}[C_1|T=0,d,q]) \\
    & \ \ \ \ \ \ \ \ \ \ \text{ with, in the general case, } \mathbb{E}[C_1|T=t,d,q] \neq 0 \ \ \forall t \\
    & \ \ \ \ \ \ \ \ \ \ \text{ and } \mathbb{E}[C_1|T=1,d,q] - \mathbb{E}[C_1|T=0,d,q] \neq 0
\end{align*}
Thereby,
\begin{enumerate}
    \item Local mediating effects are given as
    \begin{align*}
        \effect^{f^*}_{Q}(c_1,c_2,d,q) = \alpha_{QT}(q - (1-c_1)) \ \text{ and } \ \effect^{f^*}_{D}(c_1,c_2,d,q) = \frac{3\alpha_D}{10} + \alpha_{DT}(\frac{3c_1}{10} - \frac{1}{2})
    \end{align*}
    so $\mathbb{E}[\effect^{f^*}_{Q}(c_1,c_2,d,Q) | c_1, c_2] = 0$ but $\mathbb{E}[\effect^{f^*}_{D}(c_1,c_2,D,q) | c_1, c_2] \neq 0$. This illustrates the relevance of Property \ref{prop:local_mediator_effect_confounders} to isolate the fact that $Q$ is not an actual mediator conditionally on confounders.
    \item Local confounding effects are given as
    \begin{align*}
        \effect^{f^*}_{C_2}(c_1,c_2,d,q) &= \alpha_{2T}(c_2 - \frac{1}{2}) \\
        \effect^{f^*}_{C_1}(c_1,c_2,d,q) &= \alpha_1(\mathbb{E}[C_1|T=0,d,q] - \mathbb{E}[C_1|T=1,d,q]) - \alpha_{1T}\mathbb{E}[C_1|T=1,d,q] \\
    \end{align*}
    so $\mathbb{E}[\effect^{f^*}_{C_2}(C_1,C_2,d,q)] = 0$ but $\mathbb{E}[\effect^{f^*}_{C_1}(C_1,C_2,D,q)] \neq 0$. This illustrates the relevance of the two following results, which themselves generalise results of Section \ref{sec:local_bias}, to isolate the fact that $C_2$ is not an actual confounder of the relationship between treatment-outcome, treatment-mediator and mediator-outcome relationships.
    \end{enumerate}
    \begin{lemma}[Integration of the local confounding effect with mediators, true model] \label{lemma:local_confounding_effect_mediators}
 Let $M$ denote post-treatment and pre-outcome variables. Define $\mathcal{H}(C)$ as follows :
 \begin{align*}
 \mathcal{H}(C) : \ \ \forall t, m, \ Y(t,m) \indep T | C \text{ and } Y(t,m) \indep M | T, C,
 \end{align*}
 or, in other words, $C$ includes all confounders of the treatment-outcome and mediator-outcome relationships. We further assume consistency of the potential outcome, i.e. $Y(T,M) = Y$. Let $C_1, C_2$ be two pre-treatment covariates such that $\mathcal{H}(C_1, C_2)$ holds. If, additionally, $C_2$ is not a confounder, i.e.\ $\mathcal{H}(C_1)$ holds, then the integral of the local confounding effect of $f^*$ w.r.t. $C_2$ on the joint distribution of covariates for fixed values of mediators is null, i.e.
\begin{align*}
    \forall m, \ \ \mathbb{E}[\effect^{f^*}_{C_2}(C_1,C_2,m)] = 0.
\end{align*}
\end{lemma}
\begin{corollary}[Integration of the local confounding effect with mediators, black-box model] \label{cor:local_confounding_effect_blackbox_mediators}
 Let $C_1, C_2$ be two pre-treatment covariates and $M$ post-treatment and pre-outcome variables such that $C_2 \indep T, M | C_1$. Then the integral of the local confounding effect w.r.t. $C_2$ on the joint distribution of covariates is null, i.e.
\centerline{ $\forall m, \ \ \mathbb{E}[\effect^{f}_{C_2}(C_1,C_2,m)] = 0.$ }
\end{corollary}
The proofs can be found in Supplements \ref{proof:local_confounding_effect_mediators} and \ref{proof:local_confounding_effect_blackbox_mediators}

\subsection{Dependent Mediators}

We now assume the model :
\begin{align*}
    Q &\sim \text{Uniform}(0,1) \\
    T &\sim \text{Bernoulli}(0.5) \\
    D|T,Q &\sim \text{Bernoulli}(\frac{4}{5} - \frac{3}{5}\frac{T+Q}{2}) \\
    Y &= \alpha_Q Q + \alpha_D D + \alpha_T T + \alpha_{DT} DT + \alpha_{QT} QT + \epsilon, \ \ \ \ \ \mathbb{E}[\epsilon | D, Q, T] = 0.
\end{align*}

We have
\begin{align*}
    \effect^{f^*}_{T \rightarrow Y | D, Q}(d,q) &= \alpha_T + \alpha_{DT}d + \alpha_{QT}q \\
    \effect^{f^*}_{T \rightarrow Y | Q}(q)
    &= \alpha_T + \alpha_{DT}\mathbb{E}[D|T=1,q] + \alpha_{QT}q + \alpha_D(\mathbb{E}[D|T=1,q] - \mathbb{E}[D|T=0,q])   \\
    &= \alpha_T + \alpha_{DT}(\frac{1}{2} - \frac{3q}{10}) + \alpha_{QT}q - \frac{3\alpha_D}{10} \\
    \effect^{f^*}_{T \rightarrow Y | D}(d) &= \alpha_T + \alpha_Q(\mathbb{E}[Q|T=1,d] - \mathbb{E}[Q|T=0,d] ) +  \alpha_{DT}d + \alpha_{QT}\mathbb{E}[Q|T=1,d] \\
    &= \alpha_T - \frac{3\alpha_Q}{(13 - 6d)(7 + 6d)} +  \alpha_{DT}d + \alpha_{QT}\frac{7 - 4d}{13 - 6d} \\
    \effect^{f^*}_{T \rightarrow Y | \emptyset}
    &= \alpha_T + \alpha_{DT}\mathbb{E}[D|T=1] + \alpha_{QT}q + \alpha_D(\mathbb{E}[D|T=1] - \mathbb{E}[D|T=0])   \\
        &= \alpha_T + \frac{7\alpha_{DT}}{20} + \alpha_{QT}q - \frac{3\alpha_D}{10}   \\
\end{align*}
where we used
\begin{align*}
    \mathbb{E}[Q|T=1,d] &= \frac{7 - 4d}{13 -6d} \\
    \mathbb{E}[Q|T=0,d] &= \frac{4 + 2d}{7 + 6d} \\
    \mathbb{E}[Q|T=1,d] - \mathbb{E}[Q|T=0,d] &= \frac{-3}{(13 - 6d)(7 + 6d)} 
\end{align*}
which we prove from
\begin{align*}
    p(d|q,t)
    &= dp(d=1|q,t) + (1-d)p(d=0|q,t) \\
    &= d(\frac{4}{5} - \frac{3}{10}(t+q)) + (1-d)(\frac{1}{5} + \frac{3}{10}(t+q) ) \\
    p(d|t)
    &= \mathbb{E}[p(d|Q,t) | t] \\
    &= \mathbb{E}[p(d|Q,t)] \text{ as } Q \indep T \\
    &= \frac{7}{20} + \frac{3t}{10} + \frac{3d}{5}(\frac{1}{2} - t) \\
    p(q|d,t)
    &= \frac{p(d|q,t)p(q|t)p(t)}{p(d|t)p(t)} \\
    &= \frac{p(d|q,t)}{p(d|t)} \\
    &= \frac{1 + \frac{3}{2}(t+q) + 3d(1 - t - q)}{\frac{7}{4} + \frac{3t}{2} + 3d(\frac{1}{2} - t)} \\
    \mathbb{E}[Q|T=1,d]
    &= \int \frac{1 + \frac{3}{2} + \frac{3q}{2} - 3dq}{\frac{7}{4} + \frac{3}{2} - \frac{3d}{2}}q dq \\
    &= \frac{\frac{1}{2}\frac{5}{2} + \frac{1}{3}(\frac{3}{2} - 3d)}{\frac{13}{4} - \frac{3d}{2}} \\
    &= \frac{7 - 4d}{13 - 6d} \\
    \mathbb{E}[Q|T=0,d]
    &= \int \frac{1 + 3d + q(\frac{3}{2} - 3d)}{\frac{7}{4} + \frac{3d}{2}}q dq \\
    &= \frac{\frac{1}{2}(1 + 3d) + \frac{1}{3}(\frac{3}{2} - 3d)}{\frac{7}{4} + \frac{3d}{2}} \\
    &= \frac{4 + 2d}{7 + 6d}
\end{align*}
Thereby,
\begin{align*}
    \effect^{f^*}_{Q}(d,q)
    &= \effect^{f^*}_{T \rightarrow Y | D, Q}(d,q) - \effect^{f^*}_{T \rightarrow Y | D}(D) \\
    &= \frac{3\alpha_Q}{(13 - 6d)(7 + 6d)} - \alpha_{QT}\frac{7 - 4d}{13 - 6d}
\end{align*}
Notably, we note that $\mathbb{E}[\effect^{f^*}_{Q}(d,Q)] \neq 0$. Thereby, the local mediating effect as defined in Definition \ref{def:local_mediating_effect} is not able to isolate the absence of mediation from $Q$. However, defining
\begin{align} \label{eq:alternative_pathwise_effect}
    \effect^{f}_{C_i, \text{alternative}}(q) := \effect^{f}_{T \rightarrow Y | C_i}(c_i) - \effect^{f}_{T \rightarrow Y | \emptyset}
\end{align}
we note that $\mathbb{E}[\effect^{f^*}_{Q, \text{alternative}}(Q)] = 0$ and $\mathbb{E}[\effect^{f^*}_{D, \text{alternative}}(D,q)] \neq 0$. Thereby, an alternative definition of the local mediating effect, as given in \ref{eq:alternative_pathwise_effect}, is able to isolate the absence of mediation from $Q$. This actually holds in a more general setting.
\begin{property}[Ancestor of outcome] \label{prop:local_mediator_effect_alternative} Let $M$ be a post-treatment and pre-outcome variable. 
Assume that $M \indep T$, or in other words $M$ is not really a mediator. Then, 
\begin{align*}
\mathbb{E}[\effect^f_{T \rightarrow M \rightarrow Y}(M)] = 0
\end{align*}
The proof can be found in Supplement \ref{proof:local_mediator_effect_alternative}. Thereby, the original definition of the local mediating effect is not always suited for mediation analysis. However, picking up the alternative definition given above will mean we will not be able to use the same quantity for mediation, bias analysis and mediation analysis. Solving this dilemma is left for future work.

\end{property}

\section{Generalisation to a Path of Length 3 or More}
\label{sec:generalisation}

Assume the path $p$ of length $L(p)+1$ is defined as $T \rightleftarrows C_{p(1)} \rightleftarrows C_{p(2)} \rightleftarrows \cdots \rightleftarrows C_{p(L(p))} \rightarrow Y$ where $\rightleftarrows $ is either $\leftarrow$ or $\rightarrow$ and there are no other paths from $T$ into any of $C_{p(1)}, \cdots, C_{p(L(p))}$ or from any of $C_{p(1)}, \cdots, C_{p(L(p))}$ into $Y$. Then the path-wise Shapley effect with respect to $p$ is 
\begin{align}
\label{eq:generalisation}
\effect^f_p(c) = \effect^f_{ T \rightarrow Y | C_{S^*}}(c) - \effect^f_{ T \rightarrow Y | C_{S^*\backslash \{p(1), ..., p(L(p))\} }}(c_{S^*\backslash \{p(1), ..., p(L(p))\}})
\end{align}
where the second term takes the coalition with all covariates in the path removed. Local confounding and moderating effects (Definitions \ref{def:local_confounding_effect} and \ref{def:local_moderating_effect}, respectively) can be generalized to paths $p$ such that $T \leftarrow C_{p(1)}$, and local mediating effects (Definition \ref{def:local_mediating_effect}) to paths $p$ such that $T \rightarrow C_{p(1)}$.
Lemma \ref{lemma:local_confounding_effect} and Property \ref{prop:local_mediator_effect_confounders} can be generalized with these effects, by replacing covariates in the conditional independence statements with all covariates in either path.

If there are other paths of the form $T \rightarrow C_{p(k)}$ or $C_{p(k)} \rightarrow Y$, then the effect from Equation \ref{eq:generalisation} represents the effect of both $p$ and the paths $T \rightleftarrows C_{p(1)} \rightleftarrows C_{p(2)} \rightleftarrows \cdots \rightleftarrows C_{p(k)} \rightarrow Y$ for all such $k$, as by grouping $C_{p(1)}, \cdots, C_{p(L(p))}$ into $C_{\{p(1), ..., p(L(p))\}}$, all these paths would be merged into a single path $T \rightleftarrows C_{\{p(1), ..., p(L(p))\}} \rightarrow Y$.

\section{Further Experimental Results and Details}
\label{sec:exp_details}

The code used for experiments is available at \url{https://github.com/oscarclivio/pwshap}.

\subsection{Details of Experiments on Synthetic Datasets}

\textbf{Models :} We model the outcome and propensity models using linear and logistic regression, respectively. Conditional distributions for PWSHAP are inferred by training an iterative imputer. Linear regressions with second order polynomial features are used to infer outcome models and logistic regressions for propensity models.

\textbf{Causal Shapley :} Regarding Causal Shapley, we model the $ \mathbb{E}\left[f\left(X_{\bar{S}}, x_{S \cup \{j\}}\right) \mid d o\left(X_{S}=x_{S}\right)\right]$ term that is added and subtracted to obtain the direct and indirect effects (see Supplement \ref{subsec:causal_shap}) by using an iterative imputer on the dataset obtained by removing the treatment column from the original dataset. $\text{do}-$distributions are modelled differently depending on the situation. They involve making variables independent from others. This is made by reshuffling the columns of these variables in the train set, then training the imputer on this modified dataset again.

\textbf{Datasets :} datasets are generated according to the models in Supplement \ref{subsec:local_bias} for local bias analysis and in Supplement \ref{subsec:local_mediation} for local mediation analysis. 200 samples are generated and between training and testing sets as a 50/50 split. We change the links between treatment and covariates to model confounding and mediation, in the following way :
\begin{itemize}
    \item Local bias analysis :
    \begin{itemize}
        \item No confounders : $p(T=1|C_1,C_2) = 0.5$
        \item $C_1$ is a confounder, $C_2$ is not : $p(T=1|C_1,C_2) = C_1$
        \item $C_1$ and $C_2$ are confounders : $p(T=1|C_1,C_2) = C_1C_2$
    \end{itemize}
    \item Local mediation analysis :
    \begin{itemize}
        \item No mediators : $Q|T \sim \text{Uniform}(0,1), D|T \sim \text{Binomial}(0.5)$.
        \item $D$ is a mediator but not $Q$ : $Q|T \sim \text{Uniform}(0,1), D|T \sim \text{Binomial}(\frac{4}{5} - \frac{3}{5}T)$
        \item $Q$ and $D$ are mediators : $D|T \sim \text{Binomial}(\frac{4}{5} - \frac{3}{5}T)$ and $Q = \frac{3}{5} \cdot T \cdot U + (1-T) \cdot (\frac{3}{5} \cdot U + \frac{2}{5})$ where $U \sim \text{Uniform}(0,1)$
    \end{itemize}
\end{itemize}

\subsection{Further Results on the UCI Dataset}

\begin{table}
\centering
\caption{Results for additional experiment on Census Income dataset where treatment is the individual's occupation. Note that $\effect^{f}_{\subalign{&\text{Occ} \leftarrow \text{Cntr} \rightarrow  \text{Capg} \rightarrow \text{Inc} \\ &\text{Occ} \leftarrow \text{Cntr} \rightarrow \text{ Inc}}} $ is defined according to the generalization from Section \ref{sec:generalisation}.}
\label{tab:uci_occ}
\begin{tabular}{cc|cccc}
\hline
\toprule
\multicolumn{2}{c}{Causal SHAP} & \multicolumn{4}{c}{ PWSHAP}  \\
\midrule
$\phi_{\text{direct}}$    & $\phi_{\text{indirect}}$    & $\effect^{f}_{\text{Occ} \rightarrow \text{Inc}}$  & $\effect^{f}_{\text{Occ} \leftarrow \text{Cntr} \rightarrow \text{ Inc}}$  & $\effect^{f}_{\subalign{&\text{Occ} \leftarrow \text{Cntr} \rightarrow  \text{Capg} \rightarrow \text{Inc} \\ &\text{Occ} \leftarrow \text{Cntr} \rightarrow \text{ Inc}}} $  & $\effect^{f}_{ \text{Relation:} \text{Occ} \rightarrow \text{Inc}}$ \\
\small 0.132 (0.22)             & \small 0  (0.027)             & \small 0.152 (1.98)     & \small 0.083     & \small 0.147 (1.13)    & \small 0.217  (0.831)  \\ \bottomrule
\end{tabular}%

\end{table}

In an additional experiment, we consider the treatment of interest to the occupation of the individual. We focus on the question: \textit{``Under what mechanisms did having a managerial occupation impact the model prediction?''}. Results comparing PWSHAP with Causal Shapley are shown in Table \ref{tab:uci_occ}. Occupation seems to have a considerable impact throughout the cohort, with the base treatment effect showing that having a managerial job increases the predicted probability of high income by 15.2 points. Moderation by relationship status had a predominant effect in the model prediction for our individual. Being unmarried has increased the positive effect of having a managerial occupation by another 21.7 points. CausalSHAP does not capture this phenomenon as all the effect of occupation is deemed direct by definition. This further shows the high resolution of PWSHAP and impossibility of explaining. The local confounding effect of the pathway through country and capital gain also had an important impact.

\subsection{Experimental Details on UCI}

\paragraph{UCI dataset}
The UCI dataset --also known as "Census Income" dataset-- predicts whether income exceeds \$50K/yr based on census data. It includes 11 features and 32,561 individuals. We select a random subsample of 5000 individuals and only consider the following 7 features: age, capital gain, native country,  income, marital status, race and relationship. The occupation (Occ) and race variables were dichotomised, respectively into managerial/non-managerial and white/non-white. Other categorical features, namely native-country, marital status and relationship were encoded into numerical values. The URL for this dataset is \url{https://archive.ics.uci.edu/ml/datasets/adult}.

\paragraph{Pre-processing and model performance}
We use a Random Forest with 500 estimators and balanced class weights for all three models: the outcome model, the race propensity score model, the occupation propensity score model. We further used a Bayesian Ridge to learn the joint distribution on all covariates. Note that we could have inferred a propensity score model from the Bayesian Ridge but decided to fit a separate model so both weights and outcome models would be modelled with a classification tree. \newline
The 5000 observations in the set were subsampled again 10 times, by taking out 20\% of the sample. Each time, we use the subsample as both a training and testing set for our models. We further use it as a reference population. The goal is to explain how the model learned on this set, therefore using the same training set for testing isn't problematic. We ultimately want a model that has high accuracy whilst limiting the compute time. Means and standard deviations over all 10 subsamples are reported.

The accuracy of our outcome model is 0.827 and the AUC 0.944. \newline
The accuracy of our race propensity score model is 0.894 and the AUC 0.950. \newline
The accuracy of our occupation propensity score model is 0.757 and the AUC 0.863.

\paragraph{Computation and packages}
Random Forest and Bayesian Ridge were implemented using the \verb sklearn  package. We use the \verb dataset_fetcher  function from the Explanation  GAME \citeAppendix{merrick2020explanation}, available at \url{https://github.com/fiddler-labs/ the-explanation-game-supplemental} (no license). Experiments were ran using a 2,6 GHz 6-Core Intel Core i7. The amount of compute time was approximately 2,500 clock-time seconds.

\subsection{Experiments on the German Credit Dataset}

\begin{figure}
\centering
\scalebox{0.8}{
\begin{tikzpicture}[node distance=15mm, >=stealth]
 \pgfsetarrows{latex-latex};
\begin{scope}
\node[rve] (Y) {Good customer};
\node[rve, above left of=Y, xshift=-2cm] (A) {Age};
\node[rve, above of=Y] (C) {Credit amount};
\node[rve, above right of=Y, xshift=2cm] (D)  {Loan Duration};
\node[rve, above  of=A] (T) {Gender};
\draw[deg] (D) to (Y);
\draw[deg] (C) to (Y);
\draw[deg] (A) to (Y);
\draw[deg] (T) to (Y);
\draw[deg] (T) to (A);
\draw[deg] (T) to (C);
\draw[deg] (A) to (C);
\draw[deg] (C) to (D);

\end{scope}
\end{tikzpicture}
}
\caption{DAG of the German Credit Dataset.}
\label{fig:dag_gcd}
\end{figure}
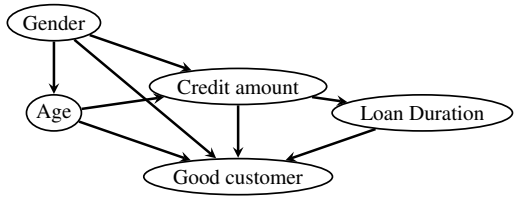

Here, we apply PWSHAP to a local mediation analysis example on the German Credit Dataset \citeAppendix{karimi2021algorithmic}. This dataset assigns people described by age and gender (1 for male, 0 for female) and seeking a loan with a certain credit amount and a certain duration to a label about whether they are good customers (1 for yes, 0 for no). The DAG is given in Figure \ref{fig:dag_gcd}.

We consider a male candidate, aged 42, asking for a loan with credit amount 5507 euros and duration 24 years. This individual's loan application gets a probability of 0.713 from the black-box model, which is below the average probability of 0.728 in the training set.
Results are shown in Figure \ref{tab:results_gdp}. PWSHAP suggests that although the total effect of gender on decisions is slightly positive, the effect of gender on decisions through credit amount, or through credit amount and loan duration, is negative. This suggests that although being male might give a slight advantage to this individual, it is compensated by high credit amount and loan duration, which are in the 65-th and 81-th percentiles, respectively, producing a relatively bad prediction compared to the average. Causal Shapley notes both a positive direct effect and a small positive indirect effect of gender on credit amount, which is not consistent with the finding that the predicted probability of being a good customer is lower than average for this individual.

\begin{table}
\centering
  \caption{Results on the German Credit Dataset. Note that the last PWSHAP effect follows the generalisation from Appendix \ref{sec:generalisation}.
  }%
\label{tab:results_gdp}
\begin{tabular}{lll}
\hline
\footnotesize{Causal SHAP} & \footnotesize{$\phi_{\text{direct}}$}   & \footnotesize{$0.025$ } \\
            & \footnotesize{$\phi_{\text{indirect}}$}   & \footnotesize{$0.004$} \\ \hline
\footnotesize{PWSHAP}      & \footnotesize{$\effect_{\text{Gender} \xrightarrow{\text{total}} \text{Good}}$}  & \footnotesize{$0.069$} \\
            & \footnotesize{$\effect_{\text{Gender} \rightarrow  \text{Amount} \rightarrow\text{Good}}$} & \footnotesize{$-0.1819$}   \\
            & \footnotesize{$\effect_{\subalign{&\text{Gender} \rightarrow  \text{Amount} \rightarrow  \text{Duration} \rightarrow\text{Good} \\ &\text{Gender} \rightarrow  \text{Amount} \rightarrow\text{Good}}}$} & \footnotesize{$-0.2371$}  \vspace{1mm} \\ \hline
\end{tabular}
\end{table}

\subsection{Computational Complexity}

\textbf{Complexity of PWSHAP effects :} For ease of notation, assume that the size of the data and the number of Monte-Carlo samples for expectations are all of size $\mathcal{O}(n)$. We denote by $p$ the number of covariates. We also assume the black-box model and the propensity score model require $\mathcal{O}(p)$ steps to be evaluated, as e.g. in linear/logistic regression, that we have access to all conditional distributions and that sampling any feature from other features can be done in $\mathcal{O}(p)$ too (e.g. if, again, all conditional distributions are GLMs).

For any sample $i$ and any path of length $l+1$ (with the treatment, $l$ covariates and the outcome), computing the path-wise PWSHAP effect mostly requires evaluating expectations of the black-box or propensity score model, where :
\begin{itemize}
\item we take $\mathcal{O}(n)$ Monte-Carlo samples of $l$ (when the treatment is observed) or $l+1$ (when the treatment is missing) variables considered to be missing from other covariates (complexity $\mathcal{O}(nlp)$) ;
\item we evaluate and aggregate the black-box for all these imputed points (complexity $\mathcal{O}(np)$).
\end{itemize}

Thus, the PWSHAP effect on a path of size $l+1$ for one sample has a complexity $\mathcal{O}(nlp)$. Aggregated over all samples, the PWSHAP of this path is of complexity $\mathcal{O}(n^2lp)$. Aggregated over all paths for all $l$, we have a complexity of at most $\mathcal{O}(n^2p^22^p)$. However, this can be reduced to $\mathcal{O}(n^2p^2)$ if the number of paths is small.

All of this assumes that we have access to conditional distributions. In practice, we do not and resort to an iterative imputer that relies on the MICE algorithm. Although, we could not find details on its complexity, scikit-learn documentation\footnote{\url{https://scikit-learn.org/stable/modules/generated/sklearn.impute.IterativeImputer.html}}  suggests that imputation and thus sampling from the conditional distribution can be done in polynomial factors w.r.t.~$n$ and $p$ too. If training the imputer is polynomial, then the overall complexity of computing all PWSHAP effects will remain polynomial in $n$ and $p$.

Thus, the method might not scale well with $p$, but with few paths in the DAG it would scale better than exact classical observational/marginal Shapley values whose complexities have a $2^p$ factor.

\textbf{Complexity of other Shapley value methods :} with the same assumptions, exact on-manifold and off-manifold Shapley values have a $\mathcal{O}(n^2p^22^p)$ or $\mathcal{O}(n^2p2^p)$ complexity, respectively. Although the general computational complexity of the approximation made by KernelSHAP \citeAppendix{lundberg2017unified} is not explicited in their work, we expect it to give better complexity at the cost of further approximating Shapley values compared to direct evaluation of expectations. TreeSHAP (an explanation model for tree ensemble models only) \citeAppendix{Lundberg2018ConsistentIF} reduces the $2^p$ factor by a square factor in the maximal depth of trees, however this method isn't model agnostic like most Shapley approaches, including PWSHAP.

\section{Running Example}
\label{sec:running_exp}

\begin{figure}
\centering
  \begin{minipage}{.5\textwidth}
    \centering
    \includegraphics[width=50mm]{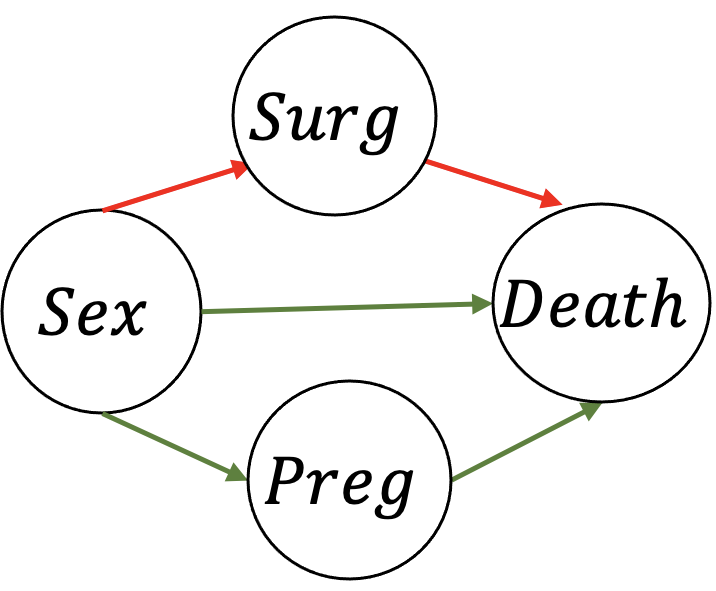}

  \end{minipage}
      \caption{DAG for the running example}
      \label{fig:dag_motivate}
  \end{figure}

Using the causal structure described in the DAG in Figure \ref{fig:dag_motivate} as a running example, we illustrate our concepts. Here, we consider three variables: \textit{Sex} (binary, denoting the female sex), \textit{Surg} (binary, denoting the execution of a surgery) and \textit{Preg} (binary, denoting whether the subject is pregnant). We denote \textit{Death} the logit of the probability of death as a result of the surgery, of its absence. We analyse the effect of the sensitive attribute \textit{Sex} on the \textit{Death} outcome, with regards to model fairness. The model is as follows:
\begin{align*}
    \mathbb{E}[\text{Sex}] &= 0.5 \\
    \mathbb{E}[\text{Preg}|\text{Sex}] &= p_\text{Preg} \cdot \text{Sex} \\
    \mathbb{E}[\text{Surg}|\text{Sex}] &=  p_\text{Surg} \cdot \text{Sex} \\
    \mathbb{E}[\text{Death}|\text{Sex, Surg, Preg}] &= f(\text{Sex, Surg, Preg}) := \alpha_\text{Sex} \text{Sex} + \alpha_\text{Surg} \text{Surg} + \alpha_\text{Preg} \text{Preg}  \\
\end{align*}

We outline a few definitions and concepts applied to this example :
\begin{itemize}
    \item On-manifold Shapley value for \text{Sex} on the empty coalition :
    \begin{align*}
        \phi^{f,\text{obs}}_{\text{Sex},\emptyset}(\text{sex}) &= (\alpha_\text{Sex} \text{sex} + \alpha_\text{Surg} \mathbb{E}[\text{Surg}|\text{sex}] + \alpha_\text{Preg} \mathbb{E}[\text{Preg}|\text{sex}])\\
        & \ \ \ \ - (\alpha_\text{Sex} \mathbb{E}[\text{Sex}] + \alpha_\text{Surg} \mathbb{E}[\text{Surg}] + \alpha_\text{Preg} \mathbb{E}[\text{Preg}]) 
    \end{align*}
    \item Off-manifold Shapley value for \text{Sex} on the empty coalition :
    \begin{align*}
        \phi^{f,\text{off-manifold}}_{\text{Sex},\emptyset}(\text{sex}) &= (\alpha_\text{Sex} \text{sex} + \alpha_\text{Surg} \mathbb{E}[\text{Surg}] + \alpha_\text{Preg} \mathbb{E}[\text{Preg}])\\
        & \ \ \ \ - (\alpha_\text{Sex} \mathbb{E}[\text{Sex}] + \alpha_\text{Surg} \mathbb{E}[\text{Surg}] + \alpha_\text{Preg} \mathbb{E}[\text{Preg}]) \\
        &= \alpha_\text{Sex} (\text{Sex} - \mathbb{E}[\text{Sex}])
    \end{align*}
    \item Causal Shapley value for \text{Sex} on the empty coalition :
    \begin{align*}
        \phi^{f,\text{causal}}_{\text{Sex},\emptyset}(\text{sex})
        &= (\alpha_\text{Sex} \text{sex} + \alpha_\text{Surg} \mathbb{E}[\text{Surg}|\text{do(Sex)}] + \alpha_\text{Preg} \mathbb{E}[\text{Preg}|\text{do(Sex)}])\\
        & \ \ \ \ - (\alpha_\text{Sex} \mathbb{E}[\text{Sex}] + \alpha_\text{Surg} \mathbb{E}[\text{Surg}] + \alpha_\text{Preg} \mathbb{E}[\text{Preg}]) \\
        &= (\alpha_\text{Sex} \text{sex} + \alpha_\text{Surg} \mathbb{E}[\text{Surg}|\text{sex}] + \alpha_\text{Preg} \mathbb{E}[\text{Preg}|\text{sex}])\\
        & \ \ \ \ - (\alpha_\text{Sex} \mathbb{E}[\text{Sex}] + \alpha_\text{Surg} \mathbb{E}[\text{Surg}] + \alpha_\text{Preg} \mathbb{E}[\text{Preg}]) \\
        &= \phi^{f,\text{obs}}_{\text{Sex},\emptyset}(\text{sex}) \ \ \ \ \text{ for this value in this specific example}.
    \end{align*}
    \item Coalition-specific Shapley effect (for \text{Sex}, our treatment of interest) on the empty coalition : one can note that the on-manifold Shapley value can be decomposed as
    \begin{align*}
        \phi^{f,\text{obs}}_{\text{Sex},\emptyset}(\text{sex}) = (\text{sex} - \mathbb{E}[\text{Sex}]) (\alpha_\text{Sex} + \alpha_\text{Preg} \cdot p_\text{Preg} + \alpha_\text{Surg} \cdot p_\text{Surg})
    \end{align*}
    where the second factor on the right-hand side is the coalition-specific Shapley effect, which here is written as
    \begin{align*}
        \effect^{f}_{\text{Sex} \rightarrow \text{Death}|\emptyset} = \mathbb{E}[f(\text{sex}=1,\text{Surg,Preg})] - \mathbb{E}[f(\text{sex}=0,\text{Surg,Preg})]
    \end{align*}
    \item Path-wise Shapley effect of \text{Surg} : here it is expressed as
    \begin{align*}
\effect^f_{\text{Surg}} &= \effect^f_{\text{Sex} \rightarrow \text{Death | Surg, Preg}}(\text{surg, preg})  - \effect^f_{\text{Sex} \rightarrow \text{Death| Preg}}(\text{preg})
\end{align*}
We can interpret as an effect through the path $\text{Sex} \rightarrow \text{Surg} \rightarrow \text{Death}$, re-noting it as $\effect^f_{\text{Sex} \rightarrow \text{Surg} \rightarrow \text{Death}}$.
\end{itemize}

\section{Proofs of Properties and Lemmas} \label{sec:proofs}

\subsection{Proof of Property \ref{prop:shapley_decomposition}}
\label{proof:shapley_decomposition}
It suffices to show that
\begin{align*}
    \coalitionshapley{f}{S}{T}{c,t} &= w^*_{S} \times \big(v_f(S \cup \{T\},c,1) - v_f(S \cup \{T\},c,0) \big)
\end{align*}
where 
\begin{align*}
     v_{f}(S \cup \{T\},c,t) &= v_{f}(S \cup \{T\},c_S,t) = \mathbb{E}_{p(C_{\bar{S}} | c_S, t)} [f(c_S, C_{\bar{s}}, t)] \\
     v_{f}(S, c) &= \mathbb{E}_{p(C_{\bar{S}}, T | c_S)} [f(c_S, C_{\bar{s}}, T)] \\
     w^*_{S} = w^*(c_S, t) &= t -  P(T=1| C_S=c_S)
\end{align*}
In other words we want to show that
\begin{align*}
    \coalitionshapley{f}{S}{T}{c,t} &= (t - p(T=1 | C_S=c_S)) \times \big(\mathbb{E}_{p(C_{\bar{S}} | c_S, 1)} [f(c_S, C_{\bar{s}}, 1)]  - \mathbb{E}_{p(C_{\bar{S}} | c_S, 0)} [f(c_S, C_{\bar{s}}, 0)]  \big)
\end{align*}

We note that
\begin{align*}
    v_f(S,c) &=  \mathbb{E}[f(c_S, C_{\bar{S}}, T) | c_S = c_S] \\
    &=  \mathbb{E} [\mathbb{E}[f(c_S, C_{\bar{S}}, T) | c_S = c_S, T] | c_S=c_S] \\
    &= p(T=t | c_S) \times \mathbb{E}[f(c_S, C_{\bar{S}}, t)|c_S=c_S, T=t]  \\ 
    & \ \ \ \ + p(T=1-t|c_S) \times \mathbb{E}[f(c_S, C_{\bar{S}}, 1-t)|c_S=c_S, T=1-t] \\
    &= p(T=t|c_S) \times v_{f}(S \cup \{T\},c,t)  \\ 
     & \ \ \ \ + p(T=1-t|c_S) \times v_{f}(S \cup \{T\},c,1-t)
\end{align*}
and, from $1=p(T=t | c_S)+p(T=1-t | c_S)$,
\begin{align*}
    &v_{f}(S \cup \{T\},c,t) - v_{f}(S,c) \\
    &= p(T=t | c_S) v_{f}(S \cup \{T\},c,t)  + p(T=1-t | c_S) v_{f}(S \cup \{T\},c,t) \\
    & \ \ \ - p(T=t | c_S) v_{f}(S \cup \{T\},c,t) - p(T=1-t | c_S) v_{f}(S \cup \{T\},c,1-t).
\end{align*}
So terms in $p(T=t | c_S)$ cancel out and we get:
\begin{align*}
    v_{f}(S \cup \{T\},c,t) - v_{f}(S,c) &=  p(T=1-t | c_S) \times [ v_{f}(S \cup \{T\},c,t)   - v_{f}(S \cup \{T\},c,1-t)] \\
    &= (t - \pi^*_S(c_S)) \times [v_{f}(S \cup \{T\},c,1)  - v_{f}(S \cup \{T\},c,0)]
\end{align*}
where $\pi^*_S(c_S) = P(T=1|C_s=c_S)$ and the second equality comes from $t \in \{0,1\}$. 

\subsection{Proof of Property \ref{prop:integration_pathwise_effect}}
\label{proof:integration_pathwise_effect}
Let $c_{-i}$ be a value of $C_{-i}$. It suffices to show that $\mathbb{E}[\effect^f_{T \rightarrow Y | C_i, C_{-i}}(C_i,c_{-i}) | C_{-i} = c_{-i}] = \effect^f_{T \rightarrow Y | C_{-i}}(c_{-i})$, which is true as
\begin{align*}
    &\mathbb{E}[\effect^f_{T \rightarrow Y | C_i, C_{-i}}(C_i,c_{-i}) | C_{-i} = c_{-i}] \\
    &= \int_{c_i} (v_f(S^* \cup \{T\}, c_i, c_{-i}, 1) - v_f(S^* \cup \{T\}, c_i, c_{-i}, 0)) \text{d}p(c_i | c_{-i}) \\
    &= \int_{c_i} (f(c_i, c_{-i}, 1) - f(c_i, c_{-i}, 0)) \text{d}p(c_i | c_{-i}) \\
    &= \int_{c_i} f(c_i, c_{-i}, 1)\text{d}p(c_i | c_{-i}) - \int_{c_i}f(c_i, c_{-i}, 0) \text{d}p(c_i | c_{-i}) \\
    &= \int_{c_i} f(c_i, c_{-i}, 1)\text{d}p(c_i | c_{-i}, t=1) - \int_{c_i}f(c_i, c_{-i}, 0) \text{d}p(c_i | c_{-i}, t=0) \text{ as } T \indep C_i | C_{-i} \\
    &= \mathbb{E}[f(C_i, c_{-i}, 1) | C_{-i}=c_{-i}, T=1] - \mathbb{E}[f(C_i, c_{-i}, 0) | C_{-i}=c_{-i}, T=0] \\
    &= v_f((S^* \backslash \{i\}) \cup \{T\}, c_{-i}, 1) - v_f((S^* \backslash \{i\}) \cup \{T\}, c_{-i}, 0) \\
    &= \effect^f_{T \rightarrow Y | C_{-i}}(c_{-i})
\end{align*}
which completes the proof.

\subsection{Proof of Property \ref{prop:error_bounds_outcome}}
\label{proof:error_bounds_outcome}
We assume that
    \begin{align*}
    \forall c, t, N, | \hat{f}_N(c,t) - f^*(c,t) | \leq e^\text{outcome}_N
    \end{align*}

Then, for any coalition $S$ without $T$, $c$ and $N$,
\begin{align*}
&| \effect^{\hat{f}_N}_{ T \rightarrow Y | C_S}(c) - \effect^{f^*}_{ T \rightarrow Y | C_S}(c)| \\
&\leq |(\mathbb{E}_{p(C_{\bar{S}}|C_S=c_s,T=1)}[\hat{f}_N(c_S,C_{\bar{S}},1)] - \mathbb{E}_{p(C_{\bar{S}}|C_S=c_s,T=0)}[\hat{f}_N(c_S,C_{\bar{S}},0)]) \\
& \ \ - (\mathbb{E}_{p(C_{\bar{S}}|C_S=c_s,T=1)}[f^*(c_S,C_{\bar{S}},1)] - \mathbb{E}_{p(C_{\bar{S}}|C_S=c_s,T=0)}[f^*(c_S,C_{\bar{S}},0)])| \\
&= |(\mathbb{E}_{p(C_{\bar{S}}|C_S=c_s,T=1)}[\hat{f}_N(c_S,C_{\bar{S}},1) - f^*(c_S,C_{\bar{S}},1)] \\
& \ \ \ \ \  - \mathbb{E}_{p(C_{\bar{S}}|C_S=c_s,T=0)}[\hat{f}_N(c_S,C_{\bar{S}},0) - f^*(c_S,C_{\bar{S}},0)])| \\
&\leq |\mathbb{E}_{p(C_{\bar{S}}|C_S=c_s,T=1)}[\hat{f}_N(c_S,C_{\bar{S}},1) - f^*(c_S,C_{\bar{S}},1)]|  \\
& \ \ \ \ \  + |\mathbb{E}_{p(C_{\bar{S}}|C_S=c_s,T=0)}[\hat{f}_N(c_S,C_{\bar{S}},0) - f^*(c_S,C_{\bar{S}},0)]| \text{ from the triangle inequality} \\
&\leq \mathbb{E}_{p(C_{\bar{S}}|C_S=c_s,T=1)}[|\hat{f}_N(c_S,C_{\bar{S}},1) - f^*(c_S,C_{\bar{S}},1)|]  \\
& \ \ \ \ \  + \mathbb{E}_{p(C_{\bar{S}}|C_S=c_s,T=0)}[|\hat{f}_N(c_S,C_{\bar{S}},0) - f^*(c_S,C_{\bar{S}},0)|] \text{ from Jensen's inequality} \\
&\leq \mathbb{E}_{p(C_{\bar{S}}|C_S=c_s,T=1)}[e^\text{outcome}_N] + \mathbb{E}_{p(C_{\bar{S}}|C_S=c_s,T=0)}[e^\text{outcome}_N] \text{ by assumption} \\
&= 2e^\text{outcome}_N
\end{align*}
which shows bound 1. Now, let $S$ as before, c, t, N,

\begin{align*}
&|\coalitionshapley{\hat{f}_N}{S}{T}{c,t}  - \coalitionshapley{f^*}{S}{T}{c,t} | \\
&= |w_S^*(t,c_S) \cdot ( \effect^{\hat{f}_N}_{ T \rightarrow Y | C_S}(c) - \effect^{f^*}_{ T \rightarrow Y | C_S}(c))| \\
&= |w_S^*(t,c_S)| | \effect^{\hat{f}_N}_{ T \rightarrow Y | C_S}(c) - \effect^{f^*}_{ T \rightarrow Y | C_S}(c)| \\
&\leq | \effect^{\hat{f}_N}_{ T \rightarrow Y | C_S}(c) - \effect^{f^*}_{ T \rightarrow Y | C_S}(c)| \text{ as } |w^*_S(t,c_S)| \leq 1 \\
&\leq  2e^\text{outcome}_N \text{ from the previous bound} \\
\end{align*}
which proves bound 2. Now, for any covariate feature $i$ as before, $c, N$
\begin{align*}
&| \effect^{\hat{f}_N}_{C_i}(c) - \effect^{f^*}_{C_i}(c) | \\
&= |\effect^{\hat{f}_N}_{ T \rightarrow Y | C_{S^*}}(c) - \effect^{\hat{f}_N}_{ T \rightarrow Y | C_{S^*\backslash \{i\} }}(c_{S^*\backslash \{i\}})   - (\effect^{f^*}_{ T \rightarrow Y | C_{S^*}}(c) - \effect^{f^*}_{ T \rightarrow Y | C_{S^*\backslash \{i\} }}(c_{S^*\backslash \{i\}}))| \\
&\leq |\effect^{\hat{f}_N}_{ T \rightarrow Y | C_{S^*}}(c) - \effect^{f^*}_{ T \rightarrow Y | C_{S^*}}(c)| \\
& \ \ \ \ + |\effect^{\hat{f}_N}_{ T \rightarrow Y | C_{S^*\backslash \{i\} }}(c_{S^*\backslash \{i\}}) - \effect^{f^*}_{ T \rightarrow Y | C_{S^*\backslash \{i\} }}(c_{S^*\backslash \{i\}})| \\
& \ \ \ \ \ \ \ \text{from the triangle inequality} \\
&\leq 2e^\text{outcome}_N + 2e^\text{outcome}_N \ \ \ \text{ from the bound on the coalition-wise Shapley effect} \\
\end{align*}
which proves bound 3.

\subsection{Proof of Property \ref{prop:error_bounds_shapley}} 
\label{proof:error_bounds_shapley}
We assume that
\begin{align*}
\forall S \text{ s.t. } T \notin S, c, N, |\estcoalitionshapley{N}{\hat{f}_N}{S}{T}{c,t} - \coalitionshapley{f^*}{S}{T}{c,t} | \leq e^\text{Shap}_N,
\end{align*}
that the arbitrary propensity score model $\pi^N$ and $\pi^*$ verify $\epsilon$-strong overlap, ie $\epsilon \leq \pi^N \leq 1 - \epsilon$, and $\epsilon \leq \pi^* \leq 1 - \epsilon$ , and that we have $\forall c, N, |\pi^N(c) - \pi^*(c) | \leq e^\text{propensity}_N$.

Then, for any $S$ s.t. $T \notin S$, we note that $\epsilon$-strong overlap for $\pi^N$ and $\pi^*$ implies $\epsilon$-strong overlap for $\pi^N_S(c_S) := \mathbb{E}_{p(C_{\bar{S}}|C_S=c_S)}[\pi^N(c_S,C_{\bar{S}})]$ and $P(T=1|C_S=c_S) = \mathbb{E}_{p(C_{\bar{S}}|C_S=c_S)}[\pi^*(c_S,C_{\bar{S}})]$ (by taking the expectation w.r.t.~$p(C_{\bar{S}} | c_S)$) and also Jensen's inequality yields
\begin{align*}
    \forall c, N, |\mathbb{E}_{p(C_{\bar{S}}|C_S=c_S)}[\pi^N(c_S,C_{\bar{S}})] -P(T=1|C_S=c_S) | \leq e^\text{propensity}_N,
\end{align*}
which further gives $\forall t, c, N, |w^N_S(c,t) - w^*_S(c,t) | \leq e^\text{propensity}_N$.

So for any $S$ s.t. $T \notin S$, $c$, $N$,
\begin{align*}
    &|w^N_S(c,0)| = |-\pi^N_S(c)| = \pi^N_S(c) \geq \epsilon \\
    &|w^N_S(c,1)| = |1-\pi^N_S(c)| = 1-\pi^N_S(c) \geq \epsilon
\end{align*}
Thereby, for any $S$ s.t. $T \notin S$, $t$, $c$, $N$,
\begin{align*}
    \frac{1}{|w^N_S(c,t)|} \leq \frac{1}{\epsilon}
\end{align*}
and, similarly,
\begin{align*}
    \frac{1}{|w^*_S(c,t)|} \leq \frac{1}{\epsilon}.
\end{align*}
We also show that $|\coalitionshapley{f^*}{S}{T}{c,t}| \leq 2||f^*||_\infty$: indeed,
\begin{align*}
    |\coalitionshapley{f^*}{S}{T}{c,t}|
    &= |w^*_S(c,t)| |\mathbb{E}_{p(C_{\bar{S}}|C_S=c_s,T=1)}[f^*(c_S,C_{\bar{S}},1)] - \mathbb{E}_{p(C_{\bar{S}}|C_S=c_s,T=0)}[f^*(c_S,C_{\bar{S}},0)]| \\
    &\leq \mathbb{E}_{p(C_{\bar{S}}|C_S=c_s,T=1)} [|f^*(c_S,C_{\bar{S}},1)|] + \mathbb{E}_{p(C_{\bar{S}}|C_S=c_s,T=0)} [|f^*(c_S,C_{\bar{S}},0)|] \\
    & \ \ \ \ \ \ \ \text{ from } |w^*_S(c,t)| \leq 1
     \text{ and the triangle inequality and Jensen's inequality} \\
     &\leq \mathbb{E}_{p(C_{\bar{S}}|C_S=c_s,T=1)} [||f^*||_\infty] + \mathbb{E}_{p(C_{\bar{S}}|C_S=c_s,T=0)} [||f^*||_\infty] \\
     &= 2 ||f^*||_\infty
\end{align*}

In the end, we have
\begin{align*}
   &|\hat{\effect}^{N,\hat{f}_N}_{ T \rightarrow Y | C_S}(c)  - \effect^{f^*}_{ T \rightarrow Y | C_S}(c)|\\
   &=  | \frac{\estcoalitionshapley{N}{\hat{f}_N}{S}{T}{c,t}}{w^N_S(c,t)} - \frac{\coalitionshapley{f^*}{S}{T}{c,t}}{w^*_S(c,t)} | \\
   &= | \frac{\estcoalitionshapley{N}{\hat{f}_N}{S}{T}{c,t} - \coalitionshapley{f^*}{S}{T}{c,t}}{w^N_S(c,t)} + \coalitionshapley{f^*}{S}{T}{c,t}(\frac{1}{w^N_S(c,t)}-\frac{1}{w^*_S(c,t)}) | \\
    &= | \frac{\estcoalitionshapley{N}{\hat{f}_N}{S}{T}{c,t} - \coalitionshapley{f^*}{S}{T}{c,t}}{w^N_S(c,t)} + \coalitionshapley{f^*}{S}{T}{c,t}\frac{w^*_S(c,t) - w^N_S(c,t)}{w^*_S(c,t) w^N_S(c,t)} | \\
    &\leq \frac{|\estcoalitionshapley{N}{\hat{f}_N}{S}{T}{c,t} - \coalitionshapley{f^*}{S}{T}{c,t}|}{|w^N_S(c,t)|} + |\coalitionshapley{f^*}{S}{T}{c,t}|\frac{|w^*_S(c,t) - w^N_S(c,t)|}{|w^*_S(c,t)| |w^N_S(c,t)|}  \ \  \ \text{from Jensen's inequality} \\
    &\leq \frac{2e^\text{Shap}_N}{\epsilon} + 2||f^*||_\infty \cdot \frac{e^\text{propensity}_N}{\epsilon^2}
\end{align*}
which proves bound 4. Bound 5 is proven similarly to bound 3 above.


\subsection{Proof of Lemma \ref{lemma:local_confounding_effect}.}

\label{proof:local_confounding_effect}
If unconfoundess w.r.t.~$C_1, C_2$ holds then
\begin{align*}
    \mathbb{E}[\effect^{f^*}_{T \rightarrow Y | C_1, C_2}(C_1, C_2)] = \mathbb{E}[\mathbb{E}[Y|T=1,C_1,C_2] - \mathbb{E}[Y|T=0,C_1,C_2]] = \text{ATE}.
\end{align*}
If unconfoundess w.r.t.~$C_1$ also holds then
\begin{align*}
    \mathbb{E}[\effect^{f^*}_{T \rightarrow Y | C_1}(C_1)] = \mathbb{E}[\mathbb{E}[Y|T=1,C_1] - \mathbb{E}[Y|T=0,C_1]] = \text{ATE}
\end{align*}
In the end,
\begin{align*}
    \mathbb{E}[\effect^{f^*}_{T \leftarrow C_2 \rightarrow Y}(C_1,C_2)] = \mathbb{E}[\effect^{f^*}_{T \rightarrow Y | C_1, C_2}(C_1, C_2) - \effect^{f^*}_{T \rightarrow Y | C_1}(C_1)] = \text{ATE} - \text{ATE} = 0.
 \end{align*}

\subsection{Proof of Property \ref{prop:local_mediator_effect_confounders}}

\label{proof:local_mediator_effect_confounders}

Let $c$ and $m_1$ be values of $C$ and $M_1$, respectively. We note that
\begin{align*}
    &\mathbb{E}[\effect^f_{T \rightarrow M_2 \rightarrow Y}(c,m_1,M_2) | C=c] \\
    &= \mathbb{E}[\effect^f_{T \rightarrow M_2 \rightarrow Y}(c,m_1,M_2) | C=c, M_1=m_1]  \text{ as } M_1 \indep M_2 | C \\
    &= 0 \ \ \ \ \ \ \ \ \ \ \ \text{ from Property \ref{prop:integration_pathwise_effect} as } M_1 \indep T | M_2, C \\
\end{align*}
which completes the proof.

\subsection{Proof of Property \ref{prop:indirect_part}}
\label{proof:indirect_part}

If a latent variable generates all pre-treatment covariates of $T$, then we can factorise the distribution of $(T,C_S,C_{\bar{S}},f\left({C}_{\bar{S}}, c_{S}, t\right))$ in the ADMG with those variables as nodes and edges $C_S \leftrightarrow C_{\bar{S}} \rightarrow f\left({C}_{\bar{S}}, c_{S}, t\right)$ and $C_{\bar{S}} \rightarrow T \gets C_S$.  We aim to apply rule 3 of Pearl's do-calculus. If we remove edges pointing into $C_S$ and $T$, we obtain an ADMG with only the edge $C_{\bar{S}} \rightarrow f\left({C}_{\bar{S}}, c_{S}, t\right)$. In this graph $T$ and $f\left({C}_{\bar{S}}, c_{S}, t\right)$ are m-separated by ${C}_{\bar{S}}$. Therefore, rule of 3 of do-calculus applies and we can remove the $T=t$ term in the do-operator of the left-hand term, yielding the right-hand term. Hence the indirect effect is zero.

\subsection{Proof of Corollary \ref{cor:local_confounding_effect_blackbox}}
\label{proof:local_confounding_effect_blackbox}
We note that
\begin{align*}
    \mathbb{E}[\effect^f_{T \leftarrow C_2 \rightarrow Y}(C_1,C_2)]
    &= \mathbb{E}[\mathbb{E}[\effect^f_{T \leftarrow C_2 \rightarrow Y}(C_1,C_2) | C_1]] \text{ from the tower property} \\
    &= \mathbb{E}[0] \text{ from Property \ref{prop:integration_pathwise_effect} as } C_2 \indep T | C_1 \\
    &= 0 \\
\end{align*}

\subsection{Proof of Lemma \ref{lemma:local_confounding_effect_mediators}.}
\label{proof:local_confounding_effect_mediators}
Let $m$ be a value of $M$. If $\mathcal{H}(C)$ holds then for any $t = 0, 1$
\begin{align*}
    \mathbb{E}[Y(t,m)]
    &= \mathbb{E}[\mathbb{E}[Y(t,m) | C]] \\
    &= \mathbb{E}[\mathbb{E}[Y(t,m) | C, t]] \text{ from } Y(t,m) \indep T | C \\
    &= \mathbb{E}[\mathbb{E}[Y(t,m) | C, t, m]] \text{ from } Y(t,m) \indep M | T, C \\
    &= \mathbb{E}[\mathbb{E}[Y | C, t, m]] \text{ from consistency.} \\
\end{align*}
so
\begin{align*}
    \mathbb{E}[\effect^{f^*}_{T \rightarrow Y | C, M}(C, m)]
    &= \mathbb{E}[\mathbb{E}[Y|C,t=1,m]] - \mathbb{E}[\mathbb{E}[Y|C,t=0,m]] \\
    &= \mathbb{E}[Y(1,m)] - \mathbb{E}[Y(0,m)] \text{ from the above} \\
    &= \text{CDE}(m).
\end{align*}
So if both $\mathcal{H}(C_1,C_2)$ and $\mathcal{H}(C_1)$ hold then
\begin{align*}
    \mathbb{E}[\effect^{f^*}_{C_2}(C_1,C_2,m)]
    &= \mathbb{E}[\effect^{f^*}_{T \rightarrow Y | C_1, C_2, M}(C_1, C_2, m) - \effect^{f^*}_{T \rightarrow Y | C_1, M}(C_1, m)] \\
    &= \text{CDE}(m) - \text{CDE}(m) \\
    &= 0. \\
 \end{align*}

\subsection{Proof of Corollary \ref{cor:local_confounding_effect_blackbox_mediators}}
\label{proof:local_confounding_effect_blackbox_mediators}
First, let's note that $C_2 \indep T, M | C_1$ implies $C_2 \indep M | C_1$ and $C_2 \indep T | C_1, M$. Let $m$ be a value of $M$. We note that
\begin{align*}
    \mathbb{E}[\effect^f_{C_2}(C_1,C_2)]
    &= \mathbb{E}[\mathbb{E}[\effect^f_{C_2}(C_1,C_2) | C_1]] \text{ from the tower property} \\
        &= \mathbb{E}[\mathbb{E}[\effect^f_{C_2}(C_1,C_2) | C_1, M]] \text{ from the property } C_2 \indep M | C_1 \\
    &= \mathbb{E}[0] \text{ from Property \ref{prop:integration_pathwise_effect} as } C_2 \indep T | C_1, M \\
    &= 0 \\
\end{align*}
which completes the proof.

\subsection{Proof of Property \ref{prop:local_mediator_effect_alternative}}
\label{proof:local_mediator_effect_alternative}
It suffices to show that $\mathbb{E}[\effect^{f}_{T \rightarrow Y | M}(M)] = \effect^{f}_{T \rightarrow Y | \emptyset}$, which is true as
\begin{align*}
    &\mathbb{E}[\effect^f_{T \rightarrow Y | M}(M)] \\
    &= \int_{m} (v_f(\{M, T\} , m, t=1) - v_f(\{M, T\} , m, t=0)) \text{d}p(m) \\
        &= \int_{m} (\mathbb{E}[f(C,m,t=1) | m, t=1] - \mathbb{E}[f(C,m,t=0) | m, t=0) \text{d}p(m) \\
    &= \int_{m} \mathbb{E}[f(C,m,t=1) | m, t=1]\text{d}p(m) - \int_{m}\mathbb{E}[f(C,m,t=0) | m, t=0]\text{d}p(m) \\
    &= \int_{m} \mathbb{E}[f(C,m,t=1) | m, t=1]\text{d}p(m|t=1) - \mathbb{E}[f(C,m,t=1) | m, t=0]\text{d}p(m|t=0) \ \ \ \text{ as } M \indep T  \\
        &= \mathbb{E}[f(C,M,t=1) | t=1] - \mathbb{E}[f(C,M,t=1) | t=0] \\
        &=  v_f(T, t=1) - v_f(T, t=0) \\
    &= \effect^{f}_{T \rightarrow Y | \emptyset} 
\end{align*}
which completes the proof.

\bibliographyAppendix{bibliography}
\bibliographystyleAppendix{apalike}

\vfill

\end{document}